\documentclass{article}

\usepackage[nonatbib,preprint]{neurips_2020}

\usepackage[utf8]{inputenc} 
\usepackage[T1]{fontenc}    
\usepackage{hyperref}       
\usepackage{url}            
\usepackage{booktabs}       
\usepackage{amsfonts}       
\usepackage{nicefrac}       
\usepackage{microtype}      
\usepackage{graphicx}
\usepackage{epsfig,color,amsmath,latexsym,amssymb,theorem}
\theoremstyle{plain}

\usepackage[linesnumbered ,ruled,vlined]{algorithm2e}
\usepackage{multirow}
\usepackage{pgffor}
\usepackage{subcaption}

\newcommand\bigzero{\makebox(0,0){\text{\Large 0}}}

\def\bX{\mathbf{X}}
\def\bx{\mathbf{x}}
\def\RR{\mathrm{\hbox{I\kern-.2em\hbox{R}}}}
\def\eps{\varepsilon}
\def\heps{\hat{\varepsilon}}
\def\var{{\rm Var}}
\def\cov{{\rm Cov}}
\def\bff{{\bf f}}

\def\sA{{\mathcal{A}}}

\def\rt{\rightarrow}

\newcommand{\ba}{\begin{array}}
\newcommand{\ea}{\end{array}}
\newcommand{\bi}{\begin{itemize}}
\newcommand{\be}{\begin{enumerate}}
\newcommand{\bc}{\begin{center}}
\newcommand{\beq}{\begin{equation}}
\newcommand{\beqa}{\begin{eqnarray}}
\newcommand{\bal}{\begin{aligned}}
\newcommand{\bbm}{\begin{bmatrix}}
\newcommand{\bcs}{\begin{cases}}
\newcommand{\ei}{\end{itemize}}
\newcommand{\ee}{\end{enumerate}}
\newcommand{\ec}{\end{center}}
\newcommand{\eeq}{\end{equation}}
\newcommand{\eeqa}{\end{eqnarray}}
\newcommand{\eal}{\end{aligned}}
\newcommand{\ebm}{\end{bmatrix}}
\newcommand{\ecs}{\end{cases}}

\title{Generalized Resubstitution for\\Classification Error Estimation}

\author{%
  Parisa Ghane \ \ \ \ \ Ulisses Braga-Neto\\
  Department of Electrical and Computer Engineering\\
  Texas A\&M University\\
  College Station, TX 77843 USA\\
  \texttt{\{pghane,ulisses\}@tamu.edu}
  }

\begin{document}

\maketitle

\begin{abstract}
We propose the family of generalized resubstitution classifier error estimators based on empirical measures. These error estimators are computationally efficient and do not require retraining of classifiers. The plain resubstitution error estimator corresponds to choosing the standard empirical measure. Other choices of empirical measure lead to bolstered, posterior-probability, Gaussian-process, and Bayesian error estimators; in addition, we propose bolstered posterior-probability error estimators as a new family of generalized resubstitution estimators. In the two-class case, we show that a generalized resubstitution estimator is consistent and asymptotically unbiased, regardless of the distribution of the features and label, if the corresponding generalized empirical measure converges uniformly to the standard empirical measure and the classification rule has finite VC dimension. A generalized resubstitution estimator typically has hyperparameters that can be tuned to control its bias and variance, which adds flexibility. Numerical experiments with various classification rules trained on synthetic data assess the finite-sample performance of several representative generalized resubstitution error estimators. In addition, results of an image classification experiment using the LeNet-5 convolutional neural network and the MNIST data set demonstrate the potential of this class of error estimators in deep learning for computer vision.
\end{abstract}
\vspace{3ex}

\section{Introduction}

Given enough training data, good classification algorithms produce
classifiers with small error rate on future data, which is also known in machine learning as the {\em generalization error}. But a
classifier is useful only if its generalization error can be stated with
confidence. Hence, at a fundamental level, one can only speak of the
goodness of a classification algorithm together with an error
estimation procedure that produces an accurate assessment of the true
generalization error of the resulting classifier.
Error estimation for classification has a long history and many different
error estimation procedures have been proposed~\cite{Tous:74,Hand:86,McLa:87,SchiHand:00, BragDougEEPR:15}. The subject has recently become a topic of concern in the deep learning community \cite{jiang2019fantastic}. Error estimators based on resampling, such as
  cross-validation \cite{LachMick:68,Cove:69,TousDona:70,Ston:74}, and
  bootstrap \cite{Efro:79,Efro:83,EfroTibs:97}, have long been popular choices of error
  estimation procedures. 
  
  However, in 
  contemporary classification applications, particularly in the case of deep learning, training can be time and resource intensive~\cite{simonyan2014very}. As a
  result, error estimators based on resampling are no longer a
  viable choice, since they require training tens or hundreds of
  classifiers on resampled versions of the training data. It has
  become instead the norm to use the test-set error, i.e., the error rate on data not used in training, to benchmark classifiers~\cite{russakovsky2015imagenet}. The test-set error estimator is an unbiased, consistent estimator of the generalization
  error regardless of the sample size or distribution of the problem~\cite{BragDougEEPR:15}. However, this is only true if the test data is truly independent of
  training, and is not reused in any way~\cite{yousefi2011multiple}. It has
  been recognized recently that this has not been always the case in image
  classification using popular benchmarks, where the same public test sets
  are heavily re-used to measure classification improvement, creating
  a situation known as ``training to the test data''
  ~\cite{recht2019imagenet}. Strictly speaking, true
  independent test sets are one-way: they can only be used once. In addition, if training and testing sample sizes are small, the test-set error estimator can display large variance, and become unreliable. All of this means that accurate test-set error estimation requires cheap access to plentiful labeled~data.
  
The alternative to resampling and test-set error estimation is
testing on the training data. The error rate on the training data is known
as the {\em resubstitution} error estimator~\cite{Smit:47}. This does not require retraining the classifier and is as fast as using a test-set error
estimator, but does not assume any separate independent test data.
The resubstitution estimator is however usually
optimisticaly biased, the more so the more the classification algorithm overfits to the training data. Optimistic bias implies that the difference between resubstitution estimate and the true error, which has been called the ``generalization gap''~\cite{keskar2016large}, is negative with a high probability. It is key therefore to investigate mechanisms to reduce the bias.

In this paper, we propose and study generalized resubstitution error estimators,
which are defined in terms of arbitrary empirical measures. In addition to plain resubstitution, this family includes well-known error estimators, such as posterior-probability \cite{LugoPawl:94}, Gaussian-process~\cite{hefny2010new}, bolstered resubstitution~\cite{BragDoug:04a}, and Bayesian~\cite{DaltDoug:11a,DaltDoug:11b} error estimators. The empirical measures used in generalized resubstitution often contain hyperparameters that can be tuned to reduce the bias and variance of the estimator with respect to plain resubstitution. 

\def\hnu{\hat{\nu}}
Working formally, given a feature vector $\bx \in R^d$, a classifier $\psi$ outputs
a label $y = \psi(\bx) \in \{0,1,\ldots,c-1\}$. A
classification rule takes sample data $S_n = \{(\bX_1,Y_1),\ldots,(\bX_n,Y_n)\} $ and produces a trained classifier
$\psi_n$. The quantity of interest is the classification error probability:
\beq
  \eps_n \,=\,\nu(\{(\bx,y): \psi_n(\bx)\neq y\})\,, 
\label{eq-err}
\eeq
where the probability measure $\nu$ is supported on $R^d \times \{0,1,\ldots,c-1\}$  and is the distribution of the pair of random variables $(\bX,Y)$.
A generalized resubstitution estimator $\hat{\eps}_n$ is defined as:
\beq
  \hat{\eps}_n\,=\, \hat{\nu}_n (\{(\bx,y): \psi_n(\bx)\neq y\})\,, 
\label{eq-grerr}
\eeq
where $\hat{\nu}_n$ is a {\em generalized empirical
probability measure}, i.e., a random probability measure supported on $R^d \times \{0,1,\ldots,c-1\}$ that is a function of the sample data.
If $\hnu_n$ is sufficiently close to $\nu$, in a suitable sense, then $\hat{\eps}_n$ is a good estimator of $\eps_n$. 

The basic example is provided by the standard empirical measure $\nu_n$ putting mass $1/n$ on each training point $(\bX_i,Y_i)$, which yields the plain resubstitution error estimator:
\beq
  \hat{\eps}^{\,r}_n\,=\, \nu_n (\{(\bx,y): \psi_n(\bx)\neq y\}) \,=\, \frac{1}{n} \sum_{i=1}^n I(\psi_n(X_i) \neq Y_i)\,,
\eeq
where $I(\cdot)$ is an indicator variable. Notice that $\nu_n$ has no hyperparameters that allow estimator bias and variance to be tuned. In the two-class case, the Vapnik-Chervonenkis (VC) theorem~\cite{DGL:96} guarantees that the empirical measure converges uniformly almost surely to the true measure, i.e., $\sup_{A \in \sA}|\nu_n(A) - \nu(A)| \rt 0$ as $n \rt \infty$, with probability 1 and regardless of $\nu$, provided that the family of sets $\sA$ is small in a precise sense. If $\sA$ is the family of sets $\{(\bx,y):\psi_n(\bx)\neq y\}$
over all possible classifiers $\psi_n$, and $\sA$ is small in the sense that  the associated classification algorithm has a finite {\em VC dimension},
then the VC Theorem implies that
$\hat{\eps}^{\,r}_n$ becomes arbitrarily close to $\eps_n$ as
$n \rt \infty$ with probability 1, in a distribution-free manner. It is a simple corollary that if, in turn, the generalized empirical measure converges uniformly almost surely to the empirical measure, i.e., $\sup_{A \in \sA}|\hnu_n(A) - \nu_n(A)| \rt 0$ as $n \rt \infty$, with probability 1 and regardless of $\nu$, then $\hat{\eps}_n$ also becomes arbitrarily close to $\eps_n$ as $n \rt \infty$, in a distribution-free manner. In other words, as sample size increases, the generalized resubstitution error estimator should look more and more like the plain resubstitution estimator.

In this paper we approach the problem of practical error estimation
from the point of view of choosing an empirical measure $\hnu_n$ and applying the corresponding generalized resubstitution estimator, while tuning the hyperparameters of $\hnu_n$ to obtain low error estimation bias and variance. 
In addition to a general theoretical framework for generalized resubstitution, our contribution includes previously unavailable
multi-class versions for some existing error estimators, and a new family of error estimators, called bolstered posterior-probability error
estimation, which is an extension of the bolstered
and posterior-probability estimators. We also discuss briefly the extension of the framework to cross-validation and test-set error estimators. Numerical experiments with synthetic data assess the finite-sample bias and variance properties of several representative generalized resubstitution error estimators using a variety of classification rules. We also present results of an image classification experiment using the LeNet-5 convolutional neural network and the MNIST data set, which display the potential of this class of error estimators in the computer vision area. 

\section{Definition and Properties}


We introduce in this section resubstitution-like error estimators based on generalized empirical measures. These error estimators share with plain resubstitution the fact that training additional classifiers is not required.
Like resubstitution, they are generally fast and low-variance. They are also nonrandomized, unless Monte-Carlo approximations are required to compute the estimator. See the Appendix for a review of the basic error estimation concepts used in the paper.


Let the feature vector $\bX \in R^d$ and the label $Y \in R$ be
jointly distributed with corresponding probability measure $\nu$, such that $\nu(R^d \times \{0,1,\ldots,c-1\}) = 1$. An event is a Borel set $A \subseteq R^d \times \{0,1,\ldots,c-1\}$. A {\em classifier} $\psi$ is a Borel-measurable function from $R^d$ to
$\{0,1,\ldots,c-1\}$. 
In practice, one collects i.i.d.\ \emph{training data} 
$S_{n}=\{(\bX_{1},Y_{1}),\ldots,$ $(\bX_{n},Y_{n})\}$, where each pair
$(\bX_i,Y_i)$ is distributed as $(\bX,Y)$, and designs a classifier $\psi_n = \Psi_n(S_n)$, by means of a classification rule $\Psi_n$. The classification error $\eps_n$ is the probability of the misclassification event $A = \{(\bx,y): \psi_n(\bx)\neq y\}$:
\beq
  \eps_n\,=\, \nu (\{(\bx,y): \psi_n(\bx)\neq y\})\,. 
\label{eq:err2}
\eeq

We define a {\em generalized empirical measure} $\hnu_n$ to be a random probability measure
supported on $R^d \times
\{0,1,\ldots,c-1\}$ almost surely that is a function of the data $S_n$.
This definition includes the standard empirical measure $\nu_n$ that puts discrete mass $1/n$ on each data point:
\beq
  \nu_n\,=\, \frac{1}{n}\sum_{i=1}^n \delta_{\bX_i,Y_i}\,,
\label{eq-empdist}
\eeq
where $\delta_{\bX_i,Y_i}$ is the (random) point measure located at $(\bX_i,Y_i)$,
defined by
\beq
  \delta_{\bX_i,Y_i}(A) \,=\, I((\bX_i,Y_i) \in A)\,,
\eeq
for each event $A$. Hence, $\nu_n(A)$ is simply the fraction 
of points in $S_n$ that are contained in $A$. Other examples of generalized empirical measures are given in Section~\ref{Sec-smooth}.

Plugging in the standard empirical measure $\nu_n$ for $\nu$ 
in (\ref{eq:err2}) yields the fraction of errors committed by
$\psi_n$ on $S_n$, i.e., the standard resubstitution error estimator:
\beq
  \hat{\eps}^{\,r}_n\,=\, \nu_n(\{(\bx,y): \psi_n(\bx)\neq y\}) \,=\, \frac{1}{n} \sum_{i=1}^n I(\psi_n(\bX_i) \neq Y_i)\,.
\label{eq-plainresub}
\eeq
By analogy, plugging in a generalized empirical measure $\hnu_n$ for $\nu$ in \eqref{eq:err2} results in a {\em generalized resubstitution error estimator}:
\begin{equation}
  \heps_n \,=\, \hnu_n(\{(\bx,y): \psi_n(\bx)\neq y\})\,.
\label{eq-gresub2}
\end{equation}
As we will see in Section~\ref{Sec-smooth}, many generalized empirical measures have tunable hyperparameters, which can be adjusted in order to reduce error estimation bias and variance.

Next, we consider the natural large-sample question of whether a generalized resubstitution error estimator approaches the true classification error as the training sample size increases to infinity. In particular, we are interested in the questions of consistency, i.e., whether $\heps_{n}\rightarrow \eps_{n}$ a.s.\ as well as asymptotic unbiasedness, i.e., whether $E[\heps_{n}] \rightarrow E[\eps_{n}]$, as $n\rightarrow \infty$. It turns out that the basic tool to address these questions is provided by the Vapnik-Chervonenkis Theorem \cite{VapnCher:71,DGL:96}.  

Note that for any fixed event $A \subseteq R^d \times \{0,1,\ldots,c-1\}$, the standard empirical measure satisfies
\beq
  \nu_n(A)\,=\, \frac{1}{n}\sum_{i=1}^n I((\bX_i,Y_i) \in A) \,\rt\, \nu(A) \:\:\: \textrm{a.s.},
\eeq
by the Strong Law of Large Numbers (SLLN). Hence, for a fixed classifier $\psi$, we can plug in the fixed set $A = \{(\bx,y): \psi(\bx)\neq y\}$ in the previous equation and conclude that the empirical classification error converges to the true error with probability 1. But this is not enough to obtain results concerning classifiers $\psi_n$ designed from the data $S_n$, since these concern events $A_n = \{(\bx,y): \psi_n(\bx)\neq y\}$, which are not fixed. What is needed instead is a {\em uniform} SLLN:
\beq
  \sup_{A \in \sA} |\nu_n(A) - \nu(A)| \,\rt\, 0\:\:\: \textrm{a.s.},
\label{eq-SLLN}
\eeq
where $\sA$ is a family of sets that must contain
all events $A_n = \{(\bx,y): \psi_n(\bx)\neq y\}$ that can be produced by the classification rule. 
In the case $c=2$, it is known that (\ref{eq-SLLN}) holds if $\sA$ is small enough, in the sense that its {\em VC dimension} $V_\sA$ is finite. The VC dimension is a nonnegative integer that measures the size of $\sA$; a smaller VC dimension implies that the classification rule is more constrained and less sensitive to the data $S_n$, i.e., it is less prone to overfitting at a fixed sample size. For example, a linear classification rule in $R^d$ has VC dimension $d+1$, which is finite, and small in low-dimensional spaces. If $V_\sA < \infty$, the {\em Vapnik-Chervonenkis Theorem}~\cite{DGL:96,Brag:20} yields the inequality:
\beq
  P\left(\sup_{A \in \sA} |\nu_n(A) - \nu(A)| > \tau
  \right) \leq 8 (n+1)^{V_\sA} e^{-n\tau^2/32},\quad \textnormal{for all } \tau > 0\,.
\label{eq-arbitC}
\eeq
The term $e^{-n\tau^2/32}$ dominates, and the bound decreases exponentially fast as $n \rt \infty$. It then follows from the First Borel-Cantelli Lemma that $\sup_{A \in \sA} |\nu_n(A) - \nu(A)| \,\rt\, 0$ a.s.\ \cite[Thm~A.8]{Brag:20}. (Strictly speaking, it is necessary to assume that events of the kind  $\sup_{A \in \sA} |\nu_n(A) - \nu(A)| > \tau$ are measurable. General conditions to ensure that are discussed in \cite{Poll:84}; such conditions are tacitly assumed throughout in this paper.)

\vspace{2ex}
\noindent
{\bf Theorem 1} {\it
In the case $c=2$, if the family $\sA$ of all events $A_n = \{(\bx,y): \psi_n(\bx)\neq y\}$ that can be produced by a classification rule has finite VC dimension, and the generalized empirical measure converges uniformly to the standard empirical measure as sample size increases, i.e.,
\beq
  \sup_{A \in \sA} |\hnu_n(A) - \nu_n(A)| \,\rt\, 0\:\:\: \textrm{a.s.},
\label{eq-conv}
\eeq
then the generalized resubstitution error estimator is consistent, $\hat{\eps}_n \rt \varepsilon _{n}$ a.s., as well as asymptotically unbiased, $E[\hat{\eps}_n] \rt E[\varepsilon_{n}]$, as $n \rt \infty$, regardless of the feature-label distribution.
} 

\vspace{1ex}
\noindent
{\bf Proof}. From 
\beq
\bal
 |\hnu_n(A) - \nu(A)| &\,=\, |\hnu_n(A) - \nu_n(A) + \nu_n(A) - \nu(A)|\\
 &\,\leq\, |\hnu_n(A) - \nu_n(A)| + |\nu_n(A) - \nu(A)|\,,
\eal
\eeq
it follows that 
\beq
  \sup_{A \in \sA} |\hnu_n(A) - \nu_n(A)| \,\leq\, \sup_{A \in \sA} |\hnu_n(A) - \nu_n(A)| + \sup_{A \in \sA} |\nu_n(A) - \nu(A)|\,.
\eeq
The first term on the right converges to zero a.s.\ by hypothesis, while the second term does so by virtue of the VC Theorem. Hence, the left-hand side must also converge to zero a.s.,
\beq
  \sup_{A \in \sA} |\hnu_n(A) - \nu(A)| \,\rt\, 0\:\:\: \textrm{a.s.}
\eeq
Since
\beq
  |\hat{\eps}_n - \varepsilon_{n}| \,=\, |\hnu_{n}(A_n) - \nu(A_n)| \,\leq\, 
\sup_{A \in \sA} |\hnu_n(A) - \nu(A)|,
\eeq
it follows that $|\hat{\eps}_n - \varepsilon_{n}| \rt 0\,$ a.s.\ and the generalized resubstitution estimator is consistent.

Furthermore, since all random variables are uniformly bounded, the Dominated Convergence Theorem implies \cite[Thm~A.7]{Brag:20} that
\beq
  |E[\hat{\eps}_n - \varepsilon_{n}]| \,\leq\, E[|\hat{\eps}_n - \varepsilon_{n}|] \,\rt\, 0\,,
\eeq
i.e., the generalized resubstitution error estimator is asymptotically unbiased. All of these results are distribution-free, holding for any feature-label distribution $\nu$. $\Box$

\vspace{2ex}
The previous result applies trivially to the plain resubstitution estimator, which has been long known, by virtue of the VC Theorem, to be consistent and asymptotically unbiased if the classification rule has finite VC dimension, regardless of the distribution of the problem \cite{DGL:96,Brag:20}. For other generalized resubstitution estimators, condition (\ref{eq-conv}) needs to be checked. The point of (\ref{eq-conv}) is that the generalized resubstitution should ``look like'' more and more to plain resubstitution as sample size increases. This makes sense, given that
plain resubstitution has very good large-sample properties in the case of finite VC dimension.

\section{Generalized Resubstitution Estimators}
\label{Sec-smooth}

We consider below several examples of generalized resubstitution estimators, which are based on a broad family of generalized empirical measures of the form:
\beq
  \hnu_n \,=\, \frac{1}{n}\sum_{i=1}^n \beta_{n,\bX_i,Y_i}\,.
\label{eq-smooth}
\eeq
where $\beta_{n,\bX_i,Y_i}$ is a random probability measure depending on the training point $\bX_i,Y_i$. Comparing this to (\ref{eq-empdist}), we realize that the generalized empirical measure in (\ref{eq-smooth}) can be seen as smoothed version of the standard empirical measure, where $\beta_{n,\bX_i,Y_i}$ provides a smoothed version of the point measure $\delta_{\bX_i,Y_i}$. This is not the only case of useful generalized empirical measure; in Section~\ref{Sec-Bayes}, we give an example that is not of the form in (\ref{eq-smooth}). 

Notice that 
a sufficient condition for (\ref{eq-conv}) in Theorem 1 is the uniform convergence of the smoothed measure $\beta_{n,\bX_i,Y_i}$ to the point measure $\delta_{\bX_i,Y_i}$ for any training point $(\bX_i,Y_i)$: 
\beq
  \sup_{A \in \sA} |\beta_{n,\bX_i,Y_i}(A) - \delta_{\bX_i,Y_i}(A)| \,\rt\, 0\:\:\: \textrm{a.s.}
\label{eq-conv2}
\eeq

\subsection{Bolstered Resubstitution}

\label{Sec-BRE}

\begin{figure}[t!]
\begin{center}
\includegraphics[width=3.5in]{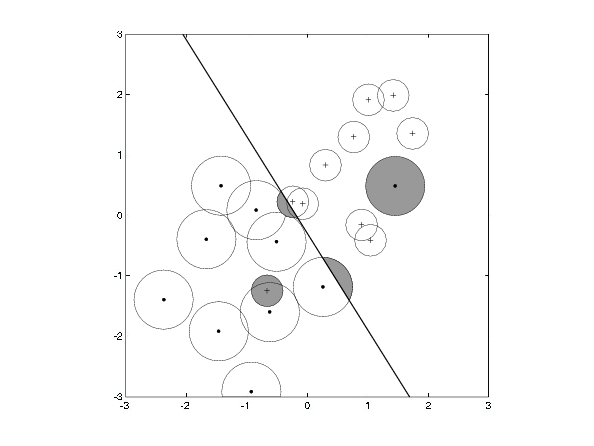}
\end{center}
\vspace*{-2ex}
\caption{Bolstered resubstitution for a linear classifier with uniform circular
bolstering kernels. The error contribution made by each point is the area of the shaded region divided by the area of the entire disk. The bolstered resubstitution error is the sum of all contributions divided by the number of points. Reproduced from \cite{BragDoug:04a}.}
\label{Fig-bolst}
\end{figure}

Given an event $A \subseteq R^d \times \{0,1,\ldots,c-1\}$, define its slices by
\beq
  A_y \,=\, \{\bx \in R^d \mid (\bx,y) \in A\}\,, \:\:\; y=0,1,\ldots,c-1\,.
\label{eq-slicey}
\eeq
It is clear that $A_y$ is an event (i.e., a Borel set) in $R^d$ for each $y$. Note that $\delta_{\bX_i,Y_i}(A) = \delta_{\bX_i}(A_{Y_i})$, where $\delta_{\bX_i}$ is a point measure in $R^d$. Similarly, let $\beta_{n,\bX_i,Y_i}(A) = \mu_{n,\bX_i,Y_i}(A_{Y_i})$, where $\mu_{n,\bX_i,Y_i}$ is an empirical measure on $R^d$.
Though discrete bolstering is possible, in practice the {\em bolstering measure} $\mu_{n,\bX_i,Y_i}$ is assumed to be absolutely continuous, with density function $p_{n,\bX_i,Y_i}(\bx)$, so that
\begin{equation*}
\beta_{n,\bX_i,Y_i}(A)\,=\, \int_{A_{Y_i}} p_{n,\bX_i,Y_i}(\bx)\,d\bx\,.
\label{eq-bpdf}
\end{equation*}
The probability densities $p_{n,\bX_i,Y_i}$ are called \emph{bolstering kernels}. 
Plugging $\beta_{n,\bX_i,Y_i}(A)$ in (\ref{eq-smooth}), and then in \eqref{eq-gresub2}, 
yields the {\em bolstered resubstitution error
estimator} proposed in \cite{BragDoug:04} (here extended to the multi-class case). Note that the misclassification event $\{(\bx,y):\psi_n(\bx) \neq y\}$ has slices $A_y = \{\bx : \psi_n(\bx) \neq 
y\}$ (If $c=2$, these reduce to the complementary decision regions $\{\bx:\psi_n(\bx) = 0\}$ and  $\{\bx:\psi_n(\bx)=1\}$.) The bolstered resubstitution error estimator can be thus written as:
\begin {equation}
\heps^{\,br}_n\,=\, \frac{1}{n} \sum_{i=1}^n \int_{\{\bx:\psi_n(\bx)\neq Y_i\}} \!\!\!\!p_{n,\bX_i,Y_i} (\bx) \, d\bx\,.
\label{eq:blstr_err}
\end{equation}
The integral in (\ref{eq:blstr_err}) gives the error contribution made
by training point $(\bX_i,Y_i)$; these are real-valued numbers between 0 and 1, unlike plain resubstitution, in which contributions are 0 or 1. See
Figure~\ref{Fig-bolst} for an illustration in the case $c=2$, where the bolstering
kernels are uniform distributions over disks centered at each of the points $\bX_i$, with radii that depend on $Y_i$. Notice that this allows counting partial errors, including errors for correctly classified
points that are near the decision boundary. 

In some cases (see an example below), it is possible to solve
the integrals in \eqref{eq:blstr_err} analytically, and the estimator is fast and low-variance. Otherwise, one has to apply approximations. For example, simple Monte-Carlo integration yields:
\begin{equation}
  \heps^{\,br}_n \,\approx\, \frac{1}{nM} \sum_{i=1}^{n}
     \sum_{j=1}^M  I(\psi_n(\bX_{ij}^{\rm MC})\neq Y_i)\,,
\label{er:blstr-mc}
\end{equation}
where $\{\bX_{ij}^{\rm MC}; j=1,\ldots ,M\}$ are random points drawn from
the density $p_{n,i}$, for $i=1,\ldots,n$. In this case, the estimation procedure is randomized due to MC sampling.

The radii of the disks in Figure~\ref{Fig-bolst} are hyperparameters that can be adjusted to control the bias of the resulting generalized resubstitution estimator. If the radii are too small, the estimator is close to plain resubstitution and could be optimistically biased; if they are too large, the estimator tends to be pessimistically biased. 


The most common choice for bolstering kernels are multivariate Gaussian densities with mean $\bX_i$ and covariance matrix $K_{n,\bX_i,Y_i}$:
\begin{equation*}
p_{n,\bX_i,Y_i}(\bx) \,=\, \frac{1}{\sqrt{(2\pi)^d \,\det(K_{n,\bX_i,Y_i})}} \exp\left(-\frac{1}{2}\, (\bx-\bX_i)^T K_{n,\bX_i,Y_i}^{-1}(\bx-\bX_i)\right)\,,
\end{equation*}
If the matrices $K_{n,\bX_i,Y_i}$ are diagonal, with freely adjustable diagonal elements, then the procedure is known as {\em Naive-Bayes bolstering}~\cite{JianBrag:14}. 

It can be shown that if $c=2$ and $\psi_n(\bx) = I(\mathbf{a}_n^T\bx +b_n>0)$ is a linear classifier, then the Gaussian-bolstered resubstitution error estimator can be computed efficiently as
\begin{equation*}
\hat{\varepsilon}_{n}^{\,\mathrm{GS}}\,=\,\frac{1}{n}\sum_{i=1}^{n}
\Phi\left(\frac{(-1)^{1-Y_i}(\mathbf{a}_n^T \bX_i + b_n)}{\sqrt{\mathbf{a}_n^TK_{n,\bX_i,Y_i} \mathbf{a}_n}}\right)\,,
\label{bolst-GresLDA}
\end{equation*}
where $\Phi(x)$ is the cumulative distribution function of a standard
$N(0,1)$ Gaussian random variable. In more general cases, one needs to employ approximations, such as Monte-Carlo sampling.

We consider below in detail the multivariate spherical Gaussian case with $K_{n,\bX_i,Y_i} = \sigma_{n,Y_i}^2I_d$. The hyperparameters here are the $c$ standard deviations $\sigma_{n,j}$ for each class (here, kernel variance is not a function of $\bX_i$). This demands much less effort than the Naive-Bayes case, which requires in general $nd$ hyperparameters. (Nevertheless, the analysis below could be extended to the Naive-Bayes case with more effort.) In \cite{BragDoug:04a} (which considers only the case $c=2$), the hyperparameters $\sigma^{n,j}$ are estimated by making the median distance of a point sampled from the corresponding kernel to the origin match the mean minimum distance $\hat{d}_{n,j}$ among training points in class $j$:
\begin{equation}
\hat{d}_{n,j} \,=\, \frac{1}{n_j} \,\sum_{i=1}^{n_j}\, ||\bX_{ij}-\bX_{ij}^\prime||\,, \quad j=0,1,\ldots,c-1\,,
\label{eq-dnj}
\end{equation}
where $n_j$ is the number of points from class $j$ ($n_j\geq 2$ is assumed), $\bX_{ij}$ is a point in class $j$, and  $\bX_{ij}^\prime$ is its nearest neighbor in class $j$. 

Now, let $R$ be the random variable corresponding to the distance to the origin of a point randomly
selected from a unit-variance spherically-symmetric density with cumulative
distribution function $F_R(x)$. The median distance of such a point to the origin is $\alpha_d = F_R^{-1}(1/2)$, where the subscript $d$ indicates explicitly that $\alpha_{d}$ depends on the dimensionality. If the density has variance $\sigma^2$, 
all distances get multiplied by $\sigma$. Hence, $\sigma_{n,j}$ is the solution of the equation 
$\sigma_{n,j} \alpha_{d} = \hat{d}_{n,j}$, i.e.,
\begin{equation}
   \sigma_{n,j} \,=\,\frac{\hat{d}_{n,j}}{\alpha_{d}}\,, \quad j=0,1,\ldots,c-1\,
\label{bolster_sigma}
\end{equation}
The constant $\alpha_{d}$ can be interpreted as a 
``dimensionality correction,'' which adjusts the value of the estimated
mean distance to account for the feature space dimensionality.
Indeed, this approach to selecting the hyperparameters is applicable to any spherically-symmetric kernel, such as the uniform disks of Figure~\ref{Fig-bolst}.
In the case of spherical Gaussian densities, $R$ is distributed as a {\em chi}
random variable with $d$ degrees of freedom, and the median $\alpha_d=F_R^{-1}(1/2)$ can be easily computed numerically. For example, the values up to five dimensions are $\alpha
_{1}=0.674$, $\alpha _{2}=1.177$, $\alpha _{3}=1.538$, $\alpha _{4}=1.832$, $\alpha _{5}=2.086$.

Next, we consider the asymptotic properties of the bolstered resubstitution estimator with spherical Gaussian kernels in the case $c=2$. First, we define a classification rule to be {\em regular} if it produces ``thin'' decision boundaries. In the general case $c \geq 2$, the decision boundary $D$ of classifier $\psi_n$ is
\beq
  D \,=\, \bigcup_{y=0}^{c-1} \partial A_y
\eeq
where $A_y = \{\bx : \psi_n(\bx) \neq y\}$ are the misclassification event slices, as defined previously, and a point is in $\partial A_y$ if it does not belong to the interior of either $A_y$ or $A_y^c$. A classification rule $\Psi_n$ is regular if $D$ has Lebesgue measure zero for all its classifiers $\psi_n$. If the distribution of $\bX$ is absolutely continuous (with respect to Lebesgue measure), i.e., if $\bX$ is a continuous feature vector in the usual sense, then the probability that a training point $\bX_i$ sits on the decision boundary is zero. The vast majority, if not all, classification rules encountered in practice are regular.



\vspace{2ex}
\noindent
{\bf Theorem 2} {\it
In the case $c=2$, if $\Psi_n$ is a regular classification rule with finite VC dimension and the distribution of $\bX$ is absolutely continuous, then the bolstered resubstitution estimator with spherical Gaussian kernels, with hyperparameters $\sigma_{n,j}$ selected as in \eqref{bolster_sigma}, is consistent and asymptotically unbiased.
} 

\def\bZ{\mathbf{Z}}
\def\bz{\mathbf{z}}
\def\bu{\mathbf{u}}
\def\bw{\mathbf{w}}

\vspace{1ex}
\noindent
{\bf Proof}.
By virtue of Theorem~1 and \eqref{eq-smooth}, it suffices to show that \eqref{eq-conv2} holds, which in the present case reduces to proving that
\beq
  \sup_{A \in \sA} |\mu_{n,\bX_i,Y_i}(A_{Y_i}) - \delta_{\bX_i}(A_{Y_i})| \,\rt\, 0\:\:\: \textrm{a.s.}\,,
\label{eq-conv2a}
\eeq
where $\sA$ s the family of all events $\{(\bx,y): \psi_n(\bx)\neq y\}$ that can be produced by the classification rule. Notice that, for any given $\tau>0$, whenever $\sup_{A \in \sA}|\mu_{n,\bX_i,Y_i}(A_{Y_i}) - \delta_{\bX_i}(A_{Y_i})| > \tau$, there is an
$A^* \in \sA$, which is a function of the data, such that $|\mu_{n,\bX_i,Y_i}(A^*_{Y_i}) - \delta_{\bX_i}(A^*_{Y_i})| > \tau$, with probability 1. In other words,
\beq
   P\left(|\mu_{n,\bX_i,Y_i}(A^*_{Y_i}) - \delta_{\bX_i}(A^*_{Y_i})| > \tau \:\bigg|\: \sup_{A \in \sA}|\mu_{n,\bX_i,Y_i}(A_{Y_i}) - \delta_{\bX_i}(A^*_{Y_i})| > \tau\right)\,=\,1\,,
\eeq
which in turn implies that
\beq
  P\left(\sup_{A \in \sA} |\mu_{n,\bX_i}(A_{Y_i}) - \delta_{\bX_i}(A_{Y_i})| > \tau\right) \,\leq\,
  P\left(|\mu_{n,\bX_i}(A^*_{Y_i}) - \delta_{\bX_i}(A^*_{Y_i})| > \tau\right) \,.
\label{eq-lemsymm1}
\eeq
By regularity of the classification rule, $\bX_i$ belongs to the interior of $A^*_{Y_i}$ or $(A^*_{Y_i})^c$ with probability 1. Hence, we can find an open ball $B(\bX_i,\rho)$ centered on $\bX_i$ that is entirely contained in $A_{Y_i}$ or $A_{Y_i}^c$. If the variance $\sigma_{n,i}^2$ tends to zero as at least $O(n)$, the Gaussian measure will concentrate exponentially fast inside such a ball, such that $P\left(|\mu_{n,\bX_i}(A^*_{Y_i}) - \delta_{\bX_i}(A^*_{Y_i})| > \tau\right) \rt 0$ exponentially fast, for any $\tau>0$, and the Theorem is proved, via (\ref{eq-lemsymm1}) and the First Borel-Cantelli Lemma.

From (\ref{eq-dnj}) and (\ref{bolster_sigma}), it suffices to show 
that the nearest neighbor $\bX_{ij}'$ to $\bX_{ij}$ converges to $\bX_{ij}$ exponentially fast as $n \rt \infty$. 
Note that, for any $\tau > 0$,\footnote{Equation \eqref{eq-nntest} appears in a similar context in the proof of the Cover-Hart Theorem for nearest-neighbor classification \cite{CoveHart:67}. The rest of the argument is distinct.} 
\beq
 P(||\bX_{ij}' - \bX_{ij}||>\tau)= P(||\bX_{kj} - \bX_{ij}|| > \tau;\,
  \textrm{ for all } k\neq i) = (1-P(||\bX_{lj} - \bX_{ij}|| < \tau))^{n_j}\!,
\label{eq-nntest}
\eeq
for some $l \neq i$. Notice that, since $P(Y = j) > 0$, $n_j \rt \infty$ as $O(n)$ a.s.\ as $n \rt \infty$. If we can show that $P(||\bX_{lj} - \bX_{ij}|| < \tau) > 0$, then it follows from (\ref{eq-nntest}) that $P(||\bX_{lj}  - \bX||>\tau) \rt 0$ exponentially fast a.s.\ and the claim is proved. To ease notation, let $\bZ'= \bX_{ij}'$ and $\bZ = \bX_{ij}$. Since $\bZ'$ and $\bZ$ are independent and identically distributed with density $p_\bX$, $\bZ' - \bZ$ has a
density $p_{\bZ'\!-\!\bZ}$, given by the classical convolution formula:
\beq
   p_{\bZ'-\bZ}(\bw)\,=\, \int p_\bX(\bw + \bu)\,p_\bX(\bw)\,d\bu\,.
\eeq
From this, we have $p_{\bZ'-\bZ}({\bf 0}) = \int p_\bX^2(\bu)\,d\bu > 0$. It follows, by continuity of the integral, that $p_{\bZ'-\bZ}$ must be nonzero in a neighborhood of ${\bf 0}$, i.e., $P(||\bZ' - \bZ|| < \tau) > 0$, as was to be shown. $\Box$

\subsection{Posterior-Probability Generalized Resubstitution}


The bolstered empirical measure relies on measures $\mu_{n,\bX_i}$ on $R^d$, which provide smoothing in the $\bX$ direction. If one performs smoothing in the $Y$ direction, the so-called posterior-probability empirical
measure results.
 
Given an event $A \subseteq R^d \times \{0,1,\ldots,c-1\}$, define the slices
\beq
  A_\bx \,=\, \{y \in \{0,1,\ldots,c-1\} \mid (\bx,y) \in A\}\,, \:\:\; \bx \in R^d\,.
\label{eq-slicex}
\eeq
(Compare to the slices in (\ref{eq-slicey}).) Note that $\delta_{\bX_i,Y_i}(A) = \delta_{Y_i}(A_{\bX_i})$, where $\delta_{Y_i}$ is a point measure on $\{0,1,\ldots,c-1\}$. Similarly, let $\beta_{n,\bX_i,Y_i}(A) = \eta_{n,\bX_i,Y_i}(A_{\bX_i})$, where $\eta_{n,\bX_i,Y_i}$ is an empirical measure on $\{0,1\ldots,c-1\}$. This is called a {\em posterior-probability measure} as $\eta_{n,\bX_i,Y_i}(A_{\bX_i})$ is to be interpreted as a ``posterior-probability'' estimate $\widehat{P}_n(Y_i\in A_{\bX_i}\mid \bX=\bX_i)$.
Plugging $\beta_{n,\bX_i,Y_i}(A)$ in (\ref{eq-smooth}), and then in \eqref{eq-gresub2}, yields the {\em posterior-probability resubstitution error
estimator} (e.g., see \cite{LugoPawl:94}, here extended to the multi-class case).

If $A = \{(\bx,y):\psi_n(\bx) \neq y\}$ is the misclassification event, then $A_\bx = \{\psi_n(\bx)\}^c$. Using the $\widehat{P}_n$ notation, it is easy to see that the posterior-probability resubstitution error estimator can be written as:
\beq
   \hat{\eps}^{\,\mathrm{ppr}}_n \,=\, \frac{1}{n} \sum_{i=1}^n 
\widehat{P}_n(\psi_n(\bX_i)\neq Y_i \mid \bX = \bX_i)\,.
\label{eq-ppr}
\eeq
Here, $\widehat{P}_n(\psi_n(\bX_i)\neq Y_i \mid \bX = \bX_i)$ is the error contribution made by training point $(\bX_i,Y_i)$, rather than 0 or 1 as in plain resubstitution. The idea is that if one is more confident that the classifier disagrees with the training label, this error
should count more, and the reverse is true if one is not. This
smoothes the error count of plain resubstitution and reduces variance.  
  

The simplest concrete example is afforded by $k$-nearest neighbor (kNN) posterior probability estimation. Let $\{y^{1}(\bx),\ldots,y^{k}(\bx)\}$ denote the labels of the $k$ nearest training points to $\bx$, for $k=1,\ldots,n$. The
$k$-nearest-neighbor (kNN) posterior probability measure is defined by 
\beq
  \widehat{P}_n(Y = y \mid \bX=\bx) \,=\, \frac{1}{k}\, \sum_{j=1}^k I(y^{j}(\bx)=y)\,,
\label{eq-pprknn}
\eeq
for $\bx \in R^d$. This makes sense since the more labels $y$ there
are in the neighborhood of $\bx$, the more likely it should be that
its label is $y$. Plugging (\ref{eq-pprknn}) into (\ref{eq-ppr}) leads to the kNN posterior-probability error estimator:
\beq
   \hat{\eps}^{\,{\rm kNN}}_n 
\,=\, \frac{1}{nk} \sum_{i=1}^n \sum_{j=1}^k I(\psi_n(\bX_i) \neq y^{j}(\bX_i))\,,
\label{eq-ppk}
\eeq
Clearly, the case $k=1$ reduces to plain
resubstitution. It is clear that if $k=k_n$ is a function of $n$, and $k_n \rt 1$ as $n\rt\infty$, (\ref{eq-conv2}) is satisfied and the kNN posterior probability estimator is consistent and asymptotically unbiased, under the conditions of Theorem~1. 

\subsection{Bolstered Posterior-Probability Generalized Resubstitution}

A novel class of generalized resubstitution estimator results if one performs smoothing in both the $\bX$ and $Y$ directions. Notice that $\delta_{\bX_i,Y_i}(A) = \delta_{\bX_i}(A_{Y_i})\delta_{Y_i}(A_{\bX_i})$, where the slices $A_y$ and $A_\bx$ are defined in (\ref{eq-slicey}) and (\ref{eq-slicex}), respectively. Let $\beta_{n,\bX_i,Y_i}(A) = \mu_{n,\bX_i,Y_i}(A_{Y_i})\eta_{n,\bX_i,Y_i}(A_{\bX_i})$, where $\mu_{n,\bX_i,Y_i}$ and $\eta_{n,\bX_i,Y_i}$ are respectively the bolstered and posterior-probability empirical measures defined previously. 
Plugging $\beta_{n,\bX_i,Y_i}(A)$ in (\ref{eq-smooth}), and then in \eqref{eq-gresub2}, yields the {\em bolstered posterior-probability resubstitution error
estimator}, a new estimator that combines features of bolstered and posterior-probability resubstitution. Using the $\widehat{P}_n$ notation, it is easy to see that the bolstered posterior-probability resubstitution error estimator can be written as:
\beq
   \hat{\eps}^{\,\mathrm{bppr}}_n \,=\, \frac{1}{n} \sum_{i=1}^n 
\left(\int_{\{\bx:\psi_n(\bx)\neq Y_i\}} \!\!\!\!\!\!\!\!p_{n,\bX_i,Y_i} (\bx)\, d\bx\right)\widehat{P}_n(\psi_n(\bX_i)\neq Y_i \mid \bX = \bX_i)\,.
\label{eq-bppr}
\eeq
This estimator seeks to combine the bias-reducing properties of the bolstered estimator with the variance-reducing properties of the posterior-probability estimator. 

For example, with $c=2$, and the Gaussian bolstering and $k$-nearest neighbor empirical measures, the bolstered posterior-probability error estimator for a linear classifier $\psi_n(\bx) = I(\mathbf{a}_n^T\bx +b_n>0)$ can be computed efficiently as
\begin{equation}
\hat{\varepsilon}_{n}^{\,\mathrm{GS-kNN}}=\frac{1}{nk} \sum_{i=1}^n \!\left[
\Phi\!\left(\frac{(-1)^{1-Y_i}(\mathbf{a}_n^T \bX_i + b_n)}{\sqrt{\mathbf{a}_n^TK_{n,\bX_i,Y_i} \mathbf{a}_n}}\right)\!\!\left(\sum_{j=1}^k 
I((-1)^{y^{j}(\bX_i)}(\mathbf{a}_n^T\bx +b_n)>0)\!\right)\!\right]\!.
\label{bolst-GresLDA2}
\end{equation}
This estimator is consistent and asymptotically unbiased under the conditions of Theorem~2 and $k_n \rt 1$ as $n\rt\infty$.

\section{Extensions}

In this section we give additional examples and discuss the extension of the framework to cross-validation and test-set error estimators. 

\subsection{Bayesian Generalized Resubstitution}
\label{Sec-Bayes}

All previous examples of generalized resubstitution were based on smoothing the error count. In this section, we given an example that shows that the family of generalized resubstitution estimators in more general than that.

Let the unknown probability measure $\nu$ belongs to a parametric family of
probability measures $\{\nu_\theta; \theta \in \Theta\}$. 
Assume a prior distribution $p(\theta)$ for the parameter, and let
$p(\theta\mid S_n)$ be its posterior distribution. We define the
{\em Bayesian empirical measure} as:
\beq
  \hnu^{\,{\rm bay}}_n(A) \,=\, \int_\Theta \nu_\theta(A)\, p(\theta\mid S_n)\,  d\theta\,,
\eeq
for each event $A \subseteq R^d \times \{0,1,\ldots,c-1\}$. Plugging this for $\hnu_n$ in \eqref{eq-gresub2} yields the {\em Bayesian generalized resubstitution estimator}.
This family of Bayesian error estimators was proposed in
\cite{DaltDoug:11a}, and later studied by the same authors in a series of papers~\cite{DaltDoug:11b,DaltDoug:11c,DaltDoug:12a,DaltDoug:12b}.
In \cite{DaltDoug:11a,DaltDoug:11b}, analytical expressions for
the Bayesian resubstitution error estimator are given in a few cases. In more
general cases, the required integrals must be
computed by numerical methods, such as Markov-Chain Monte-Carlo,
making the error estimator randomized.


\subsection{Gaussian-Process Generalized Resubstitution}

Gaussian-process classification and regression \cite{rasmussen2006gaussian} have become very popular recently. Using Gaussian process regression to estimate posterior probabilities in (\ref{eq-ppr}) leads to a {\em Gaussian-process generalized resubstitution estimator}. The idea of using Gaussian processes in a
posterior-probability error estimator was previously suggested in \cite{hefny2010new}.

Briefly, given the data $S_n = \{(\bX_1,Y_1),\ldots,(\bX_n,Y_n)\}$, consider a
Gaussian vector of {\em latent values} :
\beq
  \bff \,=\, (f_0(\bX_1),\ldots,f_0(\bX_n),\ldots,f_{c-1}(\bX_1),\ldots,f_{c-1}(\bX_n)\,,
\eeq
corresponding to samples of a vector valued Gaussian process. The 
vectors $\bff_y = (f_y(\bX_1),\ldots,f_y(\bX_n))$ are assumed to be
uncorrelated with each other, with distribution ${\cal N}({\bf 0}, K_y)$, where the covariance
matrix has elements:
\beq
  k_y(\bx,\bx')\,=\, C_y \exp\left(-\frac{1}{2\sigma_y^2}(\bx-\bx')^2\right)\,,
\label{eq-RBF}
\eeq
and the constant $C_y > 0$ ensures that $\int k_y(\bx,\bx') d\bx =
1$. Therefore $\bff$ is zero-mean and has a block-structured covariance
matrix with the $K_y$ matrices along the diagonal. The posterior
distribution of vector $\bff^* = (f^*_0(\bx),\ldots,f^*_{c-1}(\bx))$
at each test point $\bx \in R^d$ is given by~\cite{rasmussen2006gaussian}:
\beq
  p(\bff^*\mid S_n)\,=\, \int p(\bff^*\mid S_n,\bff)p(\bff \mid S_n)\, d\bff
\eeq
If the value $f^*_y(\bx)$ is large compared to the other
values $f^*_{y'}(\bx)$, for $y' \neq y$, then $\bx$ is likely to be from
class $y$, for $y=0,\ldots,c-1$. Accordingly, we define a vector
$(\xi_0(\bff^*),\ldots,\xi_{c-1}(\bff^*))$ as the output of a {\em softmax} function on $\bff^*$ and define the posterior probability function estimator
\beq
  \widehat{P}_n(Y = y \mid \bX=\bx) \,=\, \int \xi_y(\bff^*) p(\bff^*\mid S_n)\, d\bff^*\,.
\eeq
Plugging this in (\ref{eq-ppr}) yields the Gaussian-process error estimator. The
hyperparameter $\sigma_y$ correspond to the length-scale of process $f_y$,
for $y=0,\ldots,c-1$. These hyperparameters are critical to the bias properties
of the error estimator; in the literature of Gaussian processes, they are typically chosen by maximum-likelihood methods~\cite{rasmussen2006gaussian}.

\subsection{Generalized Cross-Validation and Test-Set Error Estimators}

The {\em generalized cross-validation} error estimation procedure results
from a random {\em resampling} process that produces $R$ subsets
$S^{i}_{n_i}=\{(\bX_{1}^{i},Y_{1}^{i})\,\ldots,(\bX_{n_i}^{i},Y_{n_i}^{i})\}$,
where $1 \leq n_i \leq n$. A classification
rule $\Psi_{n_i}$ is applied to $S^i_{n_i}$ to obtain a classifier
$\psi^{i}_{n_i}$, for $i = 1,\ldots,R$. It is also assumed that the resampling process
produces $R$ empirical measures $\nu_{n,1},\ldots,\nu_{n,R}$. The
generalized cross-validation error estimator is
\begin{equation*}
  \hat{\eps}_n^{\,\textrm{cvk}} \,=\, 
  \frac{1}{R}\sum_{i=1}^{R} \nu_{n,i}(\{(\bx,y): \psi^{(i)}_{n_i}(\bx)\neq y\})\,.
\label{eq-gcv}
\end{equation*}
The {\em generalized leave-one-out} error estimator corresponds to the special case $R = n$, $n_i = n-1$, and $S^{i}_{n-1}$ equal to the
original data with the point $(\bX_i,Y_i)$ deleted, for $i=1,\ldots,n$.

The {\em generalized test-set} error estimator is based
on an empirical measure $\nu^t_{m}$, which 
is a function of independent test data $S^t_m$, to produce the error estimator
\begin{equation*}
  \hat{\eps}^{\,\textrm{t}}_{n,m} \,=\, \nu^t_{m}(\{(\bx,y): \psi_n(\bx)\neq y\})\,.
\label{eq-gtest}
\end{equation*}


\section{Experimental Results}

In this section, the performance of several of the generalized resubstitution error estimators discussed in this paper is evaluated empirically, by means of classification experiments using synthetic and real data and a variety of linear and nonlinear classification rules.
Using binary synthetic data, we compare the performance of generalized resubstitution error estimators against each other and against representative cross-validation and bootstrap error estimators, both in terms of accuracy and computation speed. We also report the results of an experiment on the applicability of generalized resubstitution in image classification by convolutional neural networks (CNN), using the LeNet-5 CNN architecture and the 10-class MNIST image data set. In the latter case, due to the high-dimensionality of the feature space, we employ Naive-Bayes bolstered resubstitution, as well as a simple data-driven calibration procedure to further reduce the bias. 

\subsection{Synthetic Data Experiments}

In this section we employ synthetic data to investigate the performance of the plain resubstitution, (``standard\_resub''), bolstered resubstitution with spherical Gaussian kernels, with hyperparameter estimated as in (\ref{bolster_sigma}) (``spherical\_bolster''), the k-NN posterior-probability estimator, with $k=3$ (``3NNpp''), and the bolstered k-NN posterior probability estimator with spherical Gaussian kernels, with hyperparameters as in the previous two cases (``spherical\_bolster-3NNpp''). For comparison, we also include the 10-fold cross-validation \cite{BragDougEEPR:15} and ``zero'' bootstrap \cite{Efro:83} estimators in the experiments.

%


The generative model consists of multivariate Gaussian distributions for each of two classes, containing $d_n$ noisy features and $d-d_n$ informative features, for each sample size. The values of the noisy features are sampled independently from a zero-mean, unit-variance Gaussian distribution across both classes. 
For the informative features, the class mean vectors are $(-\delta, \dots, -\delta)$ and $(\delta, \dots, \delta)$, where the parameter $\delta>0$ is adjusted to obtain a desired level of classification difficulty. The covariance matrices for both classes are block matrices
\beq
\Sigma_{d\times d} = \sigma^2 \times \bbm
							\Sigma_{l_1\times l_1} & &  & \bigzero  \\
							&  \Sigma_{l_2\times l_2} & &  \\
							 \bigzero &  &    \ddots &  \\
							 &   & & I_{d_n \times d_n}\\
						\ebm,
\label{eq:cov}
\eeq
where $\sigma^2$ is a variance parameter and $\Sigma_{l_i}$ is an $l_i \times l_i$ matrix,
\beq
  \Sigma_{l_i\times l_i} 
   \,=\, \bbm 1 & \rho & \cdots & \rho \\
            \rho & 1 & \cdots & \rho \\
            \vdots & \vdots & \ddots & \vdots \\    
            \rho & \rho & \cdots & 1\ebm,
\eeq
representing $l_i$ correlated features, with correlation coefficient $-1<\rho<1$, such that $\sum_i l_i=d-d_n$.  

\begin{figure}
\begin{center}
\includegraphics[scale= 0.5] {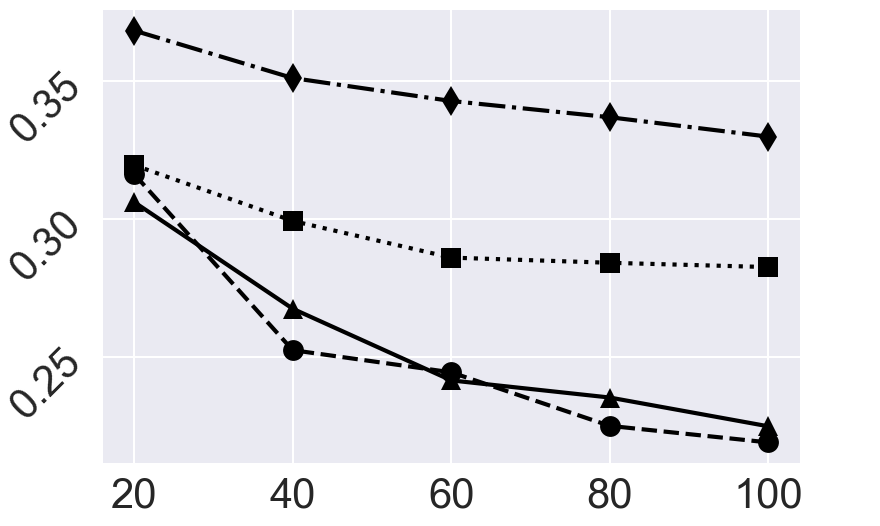}	
\includegraphics[scale= 0.4] {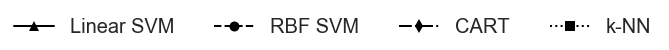}
\end{center}  
    \caption{Average classification error versus sample size.
    }
    \label{fig:true_err}
\end{figure}

In the experiments below, we considered $d=10$ features, consisting of $d_n = 4$ noisy and $d-d_n=6$ informative features. The latter are correlated in pairs, i.e., $l_1=l_2=l_3=2$, with correlation coefficient $\rho=0.2$. Four classification rules were considered: linear support vector machine (SVM), nonlinear SVM with radial basis kernel function (RBF-SVM), classification tree (CART) with stopped splitting at 5 points per leaf node, and k-nearest neighbors (k-NN), with $k=3$. We 
adjusted  $\delta$ to produce moderate classification difficulty over a range of sample sizes $n = 20,40,60,80$, and $100$; see Figure \ref{fig:true_err}. 

The true classification error was approximated using a large test data set of size 5000. The bias, variance, and RMS of each error estimator --- see equations (\ref{eq-bias}), (\ref{eq-var_est}) and (\ref{eq-err_rms}), respectively, in the Appendix --- were estimated by training each classifier 200 times using independently generated data sets and averaging. 
The bolstered resubstitution and bolstered posterior-probability error estimators for nonlinear classifiers used $M = 100$ Monte-Carlo points in their computation (the estimators are computed exactly for the linear SVM). 


\begin{figure}
\begin{center}
\begin{tabular}{cc}
\includegraphics[height=0.25\textwidth, width= 0.4\textwidth] {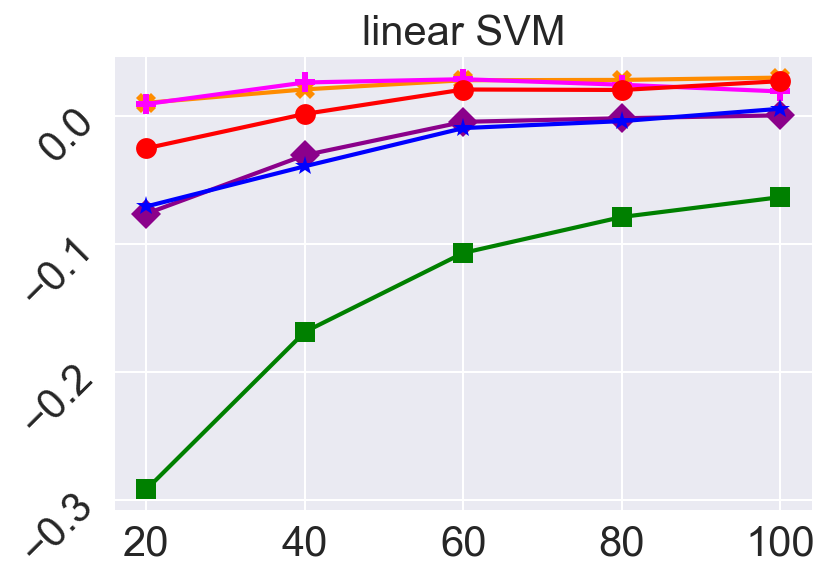}	&  
     \includegraphics[height=0.25\textwidth, width= 0.4\textwidth] {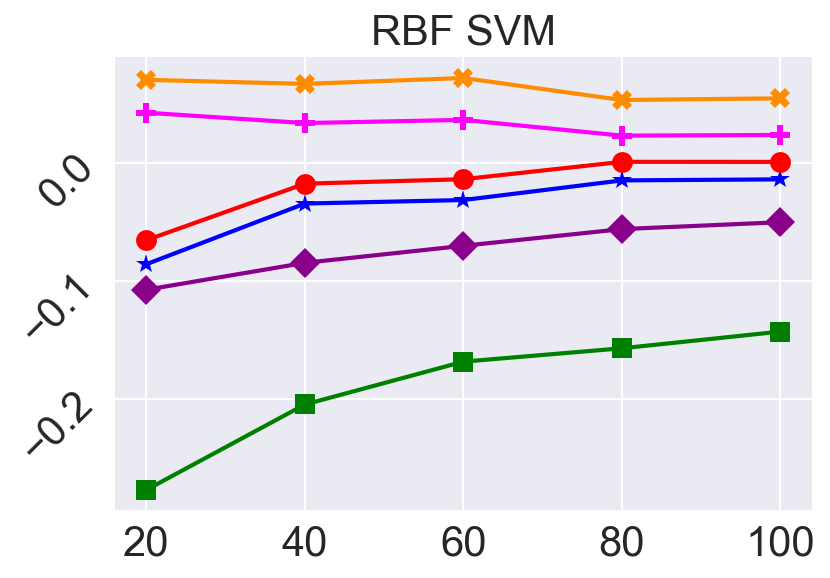} 
     \\
      \includegraphics[height=0.25\textwidth, width= 0.4\textwidth] {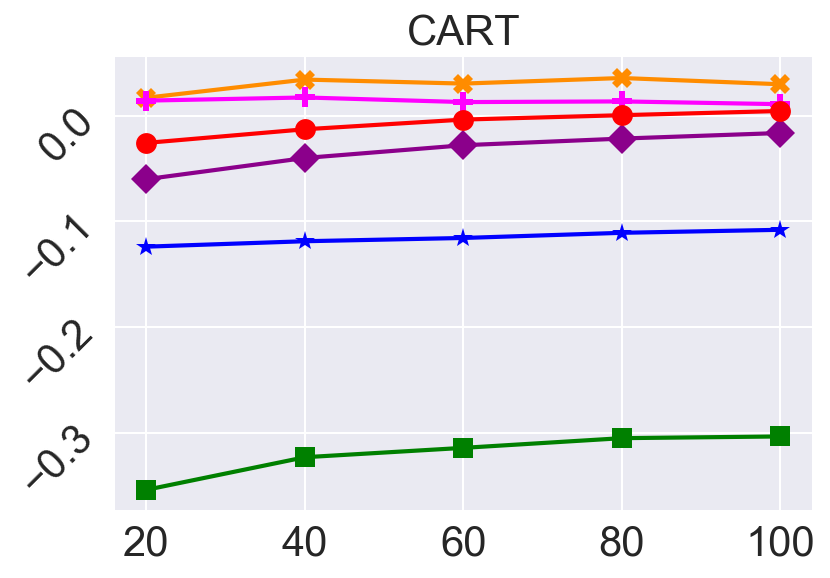} &
       \includegraphics[height=0.25\textwidth, width= 0.4\textwidth] {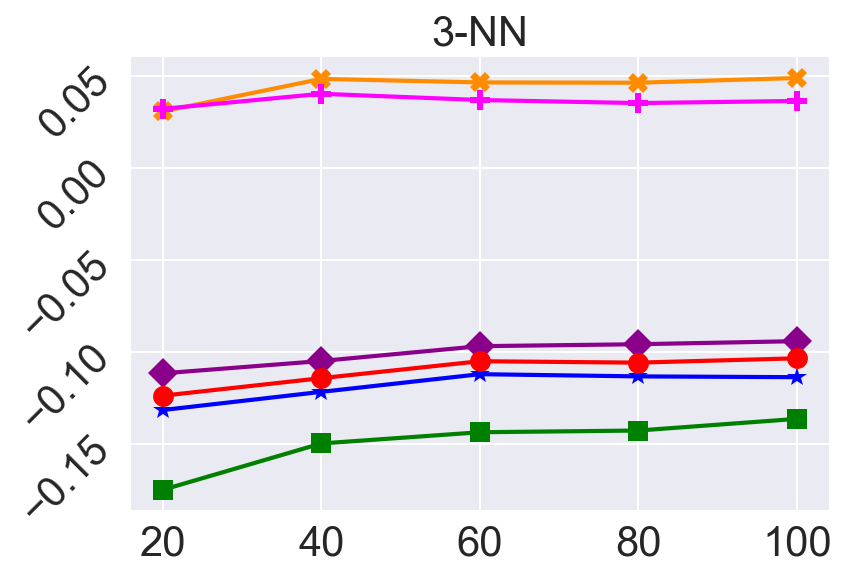}
\end{tabular}	
\includegraphics[scale= 0.4] {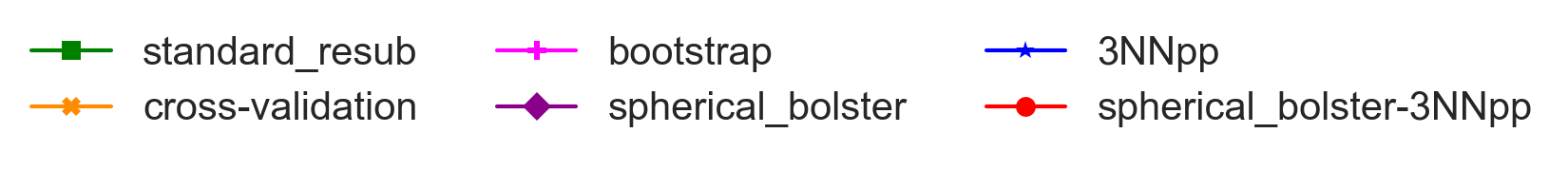}
\end{center}  
\vspace{-3ex}
    \caption{Bias versus sample size.
    }
    \label{fig:C2bias}
\end{figure}

\begin{figure}
\begin{center}
\begin{tabular}{cc}
\includegraphics[height=0.25\textwidth, width= 0.4\textwidth] {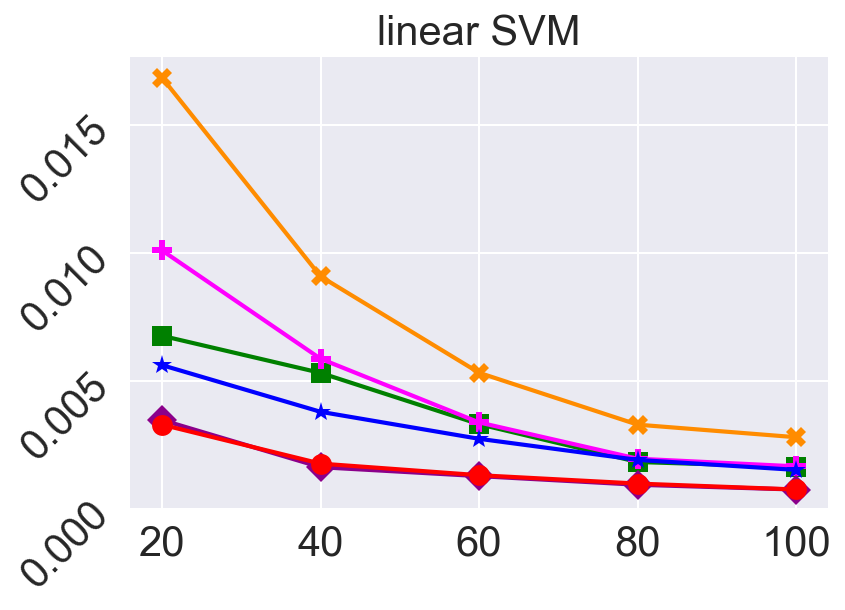}	&  
     \includegraphics[height=0.25\textwidth, width= 0.4\textwidth] {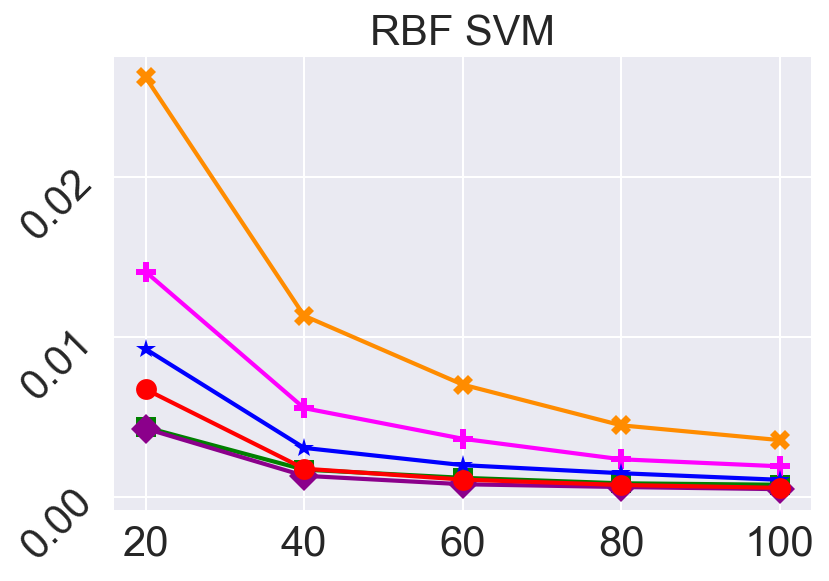} 
     \\
      \includegraphics[height=0.25\textwidth, width= 0.4\textwidth] {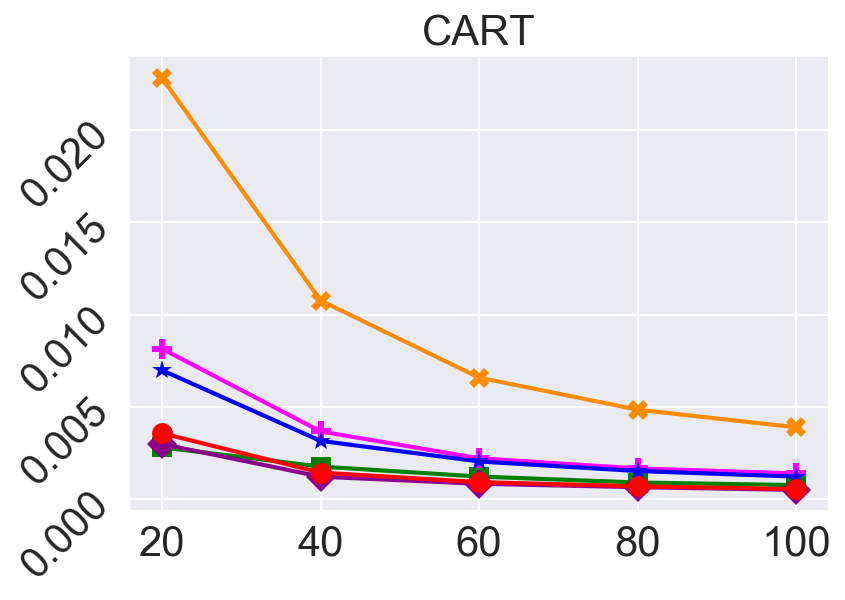} &
       \includegraphics[height=0.25\textwidth, width= 0.4\textwidth] {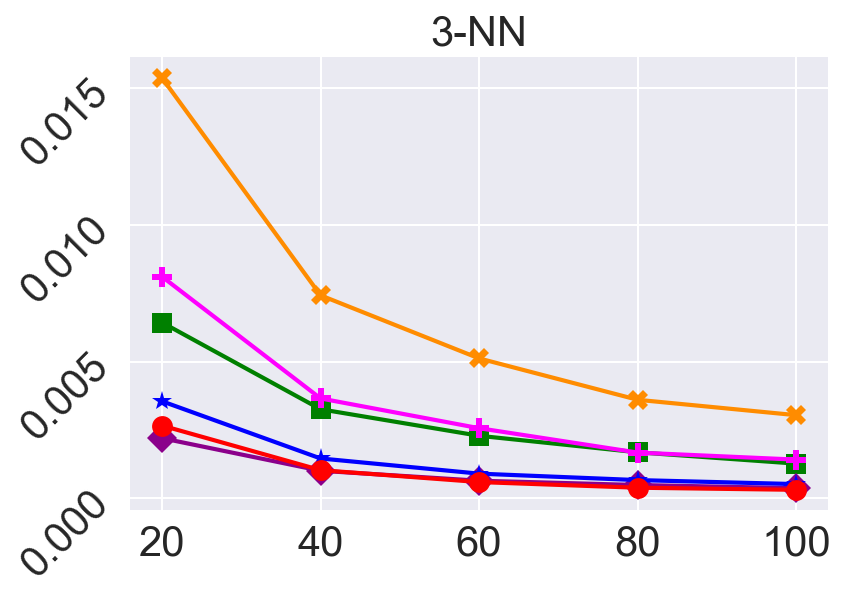}
\end{tabular}	
\includegraphics[scale= 0.4] {figs/synthetic/c2/legend.png}
\end{center}  
\vspace{-3ex}
    \caption{Variance versus sample size.
    }
    
    \label{fig:C2var}
\end{figure}

\begin{figure}
\begin{center}
\begin{tabular}{cc}
\includegraphics[height=0.25\textwidth, width= 0.4\textwidth] {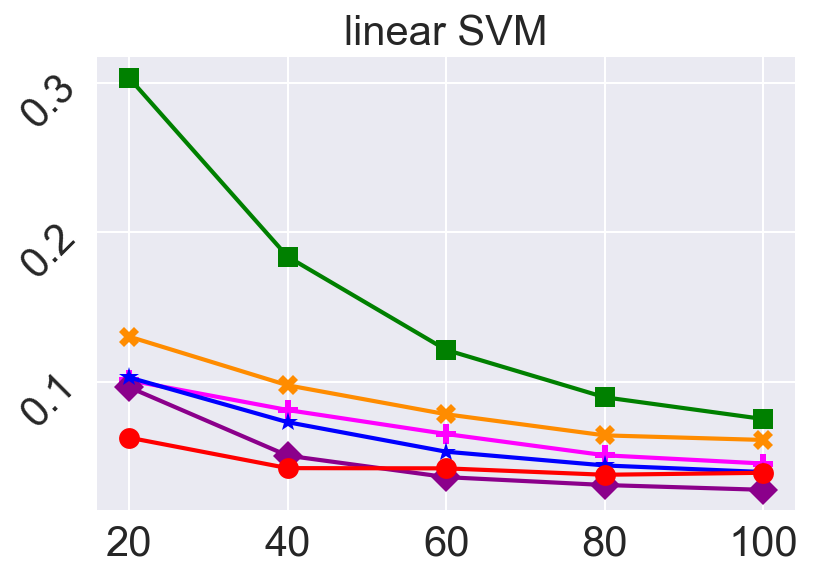}	&  
     \includegraphics[height=0.25\textwidth, width= 0.4\textwidth] {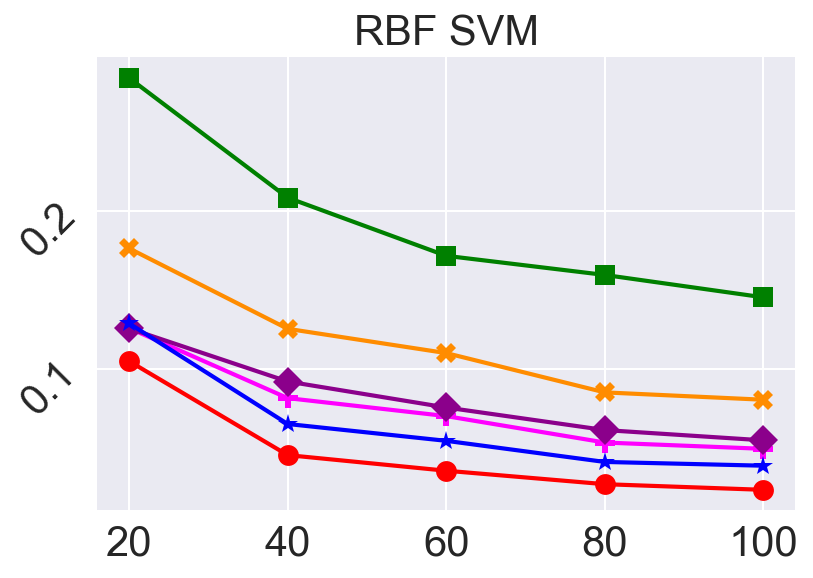} 
     \\
      \includegraphics[height=0.25\textwidth, width= 0.4\textwidth] {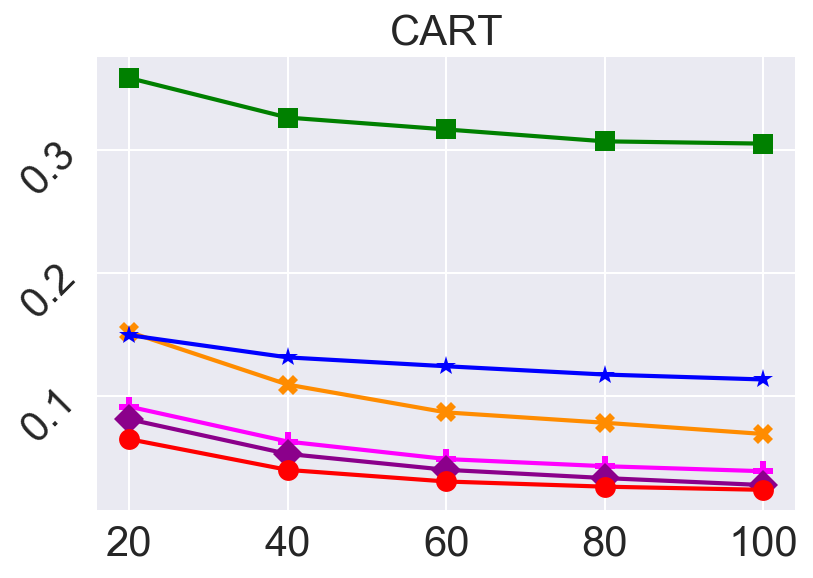} &
       \includegraphics[height=0.25\textwidth, width= 0.4\textwidth] {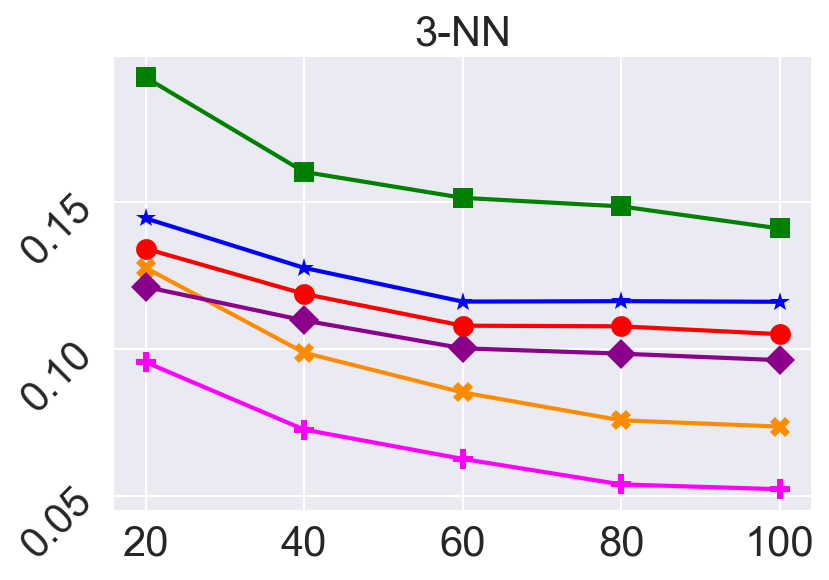}
\end{tabular}	
\includegraphics[scale= 0.4] {figs/synthetic/c2/legend.png}
\end{center}  
\vspace{-3ex}
    \caption{RMS versus sample size.
    }
    \label{fig:C2rms}
\end{figure}

The results of the experiment are displayed in Figures~\ref{fig:C2bias}, \ref{fig:C2var}, and \ref{fig:C2rms}. We can see in Figure~\ref{fig:C2bias} that plain resubstitution is heavily negatively-biased at small sample sizes, as expected. All three bolstered estimators are able to significantly reduce the bias of plain resubstitution, except in the case of 3NN classification (more on this case to follow). As sample size increases, the best estimator in terms of bias is the bolstered-3NNpp estimator, which becomes essentially unbiased, except in the case of 3NN classification. The bias properties of the bolstered estimators are best in the case of classifiers with piecewise linear boundaries (linear SVM and CART) and less so for the highly nonlinear classifiers. The 3NNpp estimator has good bias properties only in the linear SVM case (\cite{LugoPawl:94} stated that the resubstitution-like posterior-probability estimator was expected to be biased). The cross-validation and bootstrap estimators are positively biased, which is expected since both estimators employ classifiers trained on data sets of smaller effective sample size than the original one~\cite{BragDougEEPR:15}.

As seen in Figure~\ref{fig:C2var}, cross-validation is highly-variable at small sample sizes, which is also expected~\cite{BragDoug:04}. In this  figure, we can also see that all bolstered estimators display smaller variance than either cross-validation or the bootstrap, in some cases significantly so, even with the nonlinear classifiers. 

Finally, in Figure~\ref{fig:C2rms}, we can see that the RMS (which combines bias and variance in a single metric) reveals a clear superiority of the bolstered estimators over plain resubstitution, cross-validation, and bootstrap, except in the case of 3NN classification, due to the bias issue mentioned earlier. Except for the 3NN classifier, the best estimator overall is the bolstered posterior-probability error estimator. 

\begin{figure}
\begin{center}
\begin{tabular}{ccc}
\includegraphics[scale= 0.35] {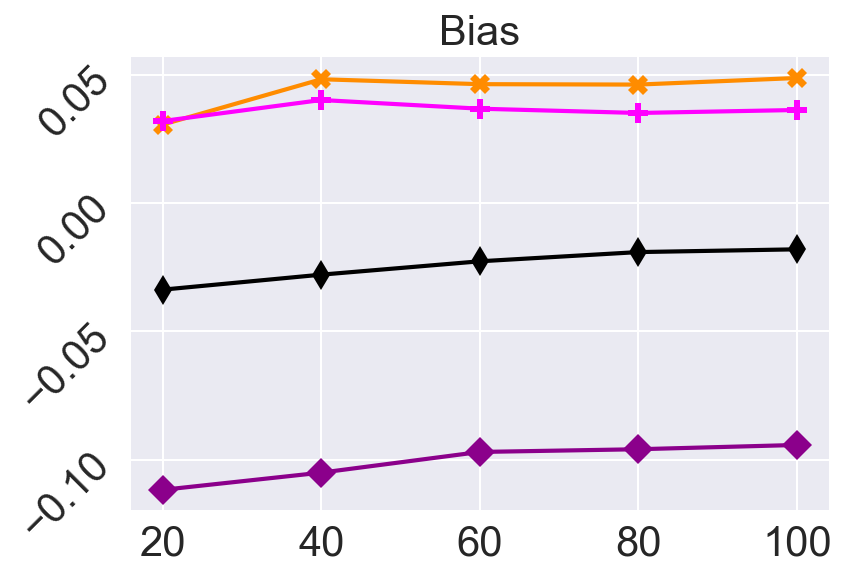}	&  
    \hspace{-0.2in}
	  \includegraphics[scale= 0.35] {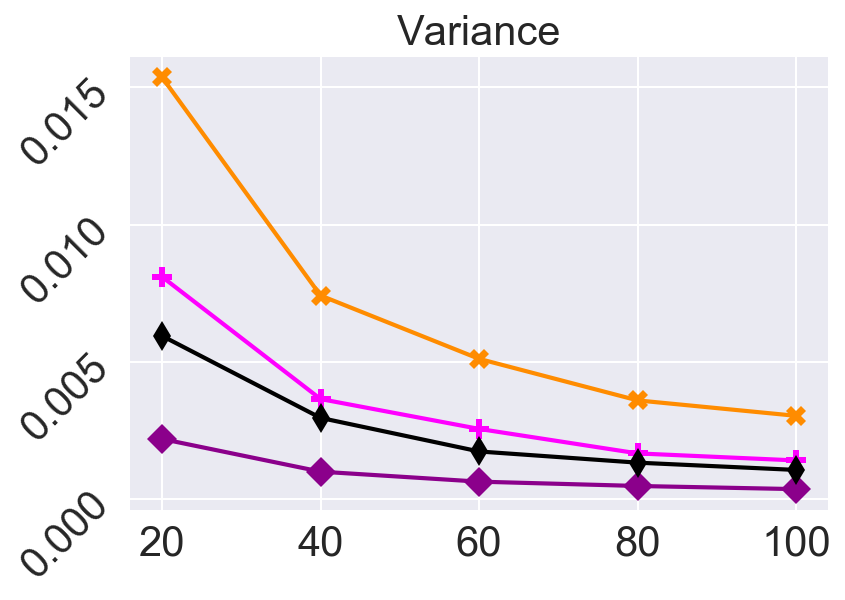} & 
	  \hspace{-0.2in}
	  \includegraphics[scale= 0.35] {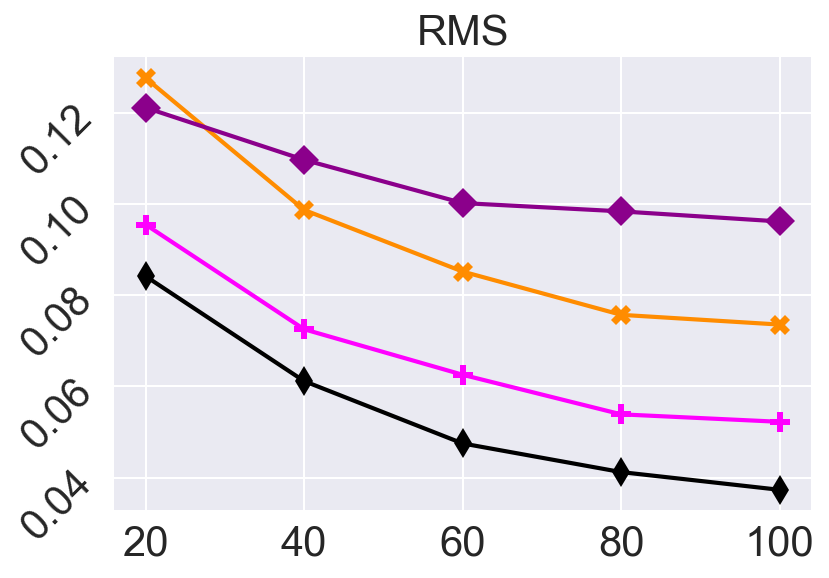} 
\end{tabular}	
\includegraphics[scale=0.4] {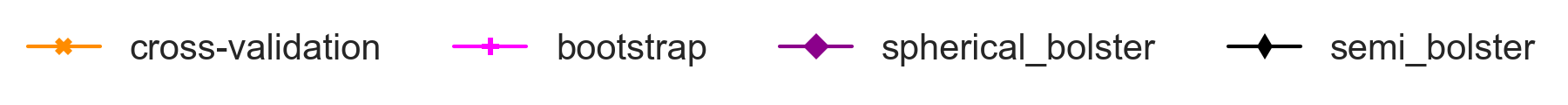}
\end{center}  
    \caption{Bias, variance, and RMS versus sample size for the 3NN classifier, including semi-bolstered resubstitution. For clarity,  only the best performing error estimators (in terms of RMS) are displayed.  
    }
    \label{fig:3NN-semibol}
\end{figure}

The difficulty of bolstering to correct the bias of resubstitution in the case of 3NN classification was already noted in \cite{braga2004bolstered} (incidentally, 3NN has infinite VC dimension and Theorem~2 does not apply). 
In that reference, the so-called ``semi-bolstered resubstitution'' error estimator, which only applies bolstering kernels to correctly-classified points, was proposed as a fix. Indeed, as can be seen in Figure \ref{fig:3NN-semibol}, using the same synthetic data as the previous experiments, semi-bolstering with spherical Gaussian kernels is capable of correcting the bias of resubstitution in the case of 3NN much more effectively than the best-RMS bolstered estimator in that case (the plain bolstered resubstitution), which coupled with a small variance, makes it the best estimator overall in terms of RMS. 

In addition to the bias-variance properties discussed above, a very important issue is the computational complexity of the various error estimators, particularly in cases where thousands (or more) error estimates must be computed, as in wrapper feature selection. Table~\ref{tab:run-time} displays the average computation time obtained by the error estimators in the experiment. We can see that the posterior-probability error estimator, despite some bias issues, is lightning fast. Combined with its small variance, this makes this estimator very attractive for computationally-expensive classification tasks. Cross-validation (at 10 folds and no repetition) is the next fastest error estimator. Its poor variance properties under small sample sizes --- and, in the case of wrapper feature selection, the issue of selection bias~\cite{AmbrMcLa:02} --- makes it unattractive. The bolstered resubstitution estimators are less fast but still much faster than the bootstrap. 

\begin{table}[]
\bc
\begin{tabular}{l|c|ccccc}
\hline
Classifier                  & $n$   & spherical & 3NNpp & spherical & cross & bootstrap \\
                  &    & bolster &  & bolster-3NNpp & validation  &  \\
\hline
\multirow{5}{*}{linear SVM} & 20  & 9.65       & 2.31  & 8.65             & 4.98             & 138.75    \\
                            & 40  & 16.66      & 1.84  & 16.48            & 6.27             & 138.83    \\
                            & 60  & 24.49      & 1.79  & 23.55            & 7.90             & 170.62    \\
                            & 80  & 33.81      & 1.77  & 33.92            & 11.34            & 234.67    \\
                            & 100 & 41.06      & 1.97  & 41.43            & 17.71            & 295.49    \\
\hline
\multirow{5}{*}{RBF SVM}    & 20  & 11.43      & 2.41  & 11.39            & 7.02             & 180.65    \\
                            & 40  & 20.87      & 2.10  & 21.38            & 8.45             & 198.10    \\
                            & 60  & 31.81      & 1.88  & 35.16            & 11.25            & 252.44    \\
                            & 80  & 45.18      & 2.01  & 55.21            & 14.70            & 327.30    \\
                            & 100 & 61.53      & 2.15  & 78.55            & 19.98            & 434.10    \\
\hline
\multirow{5}{*}{CART}       & 20  & 8.13       & 2.47  & 9.91             & 4.25             & 123.71    \\
                            & 40  & 15.00      & 2.19  & 20.45            & 5.53             & 123.13    \\
                            & 60  & 20.76      & 1.85  & 25.94            & 5.16             & 129.78    \\
                            & 80  & 28.33      & 1.79  & 32.30            & 5.70             & 133.18    \\
                            & 100 & 38.32      & 1.95  & 40.12            & 6.60             & 146.84    \\
\hline
\multirow{5}{*}{3-NN}       & 20  & 79.30      & 4.00  & 136.46           & 10.60            & 297.68    \\
                            & 40  & 137.93     & 4.01  & 152.26           & 10.22            & 282.93    \\
                            & 60  & 202.52     & 4.61  & 236.21           & 11.06            & 378.18    \\
                            & 80  & 289.80     & 6.03  & 322.32           & 13.40            & 541.11    \\
                            & 100 & 343.50     & 6.28  & 360.67           & 13.11            & 657.88   \\
\hline
\end{tabular}
\ec
\caption{Average computation time (in milliseconds). The simulations were performed on a macOS version 10.15.7, with a 3.1 GHz Dual-Core Intel Core i7 processor and 16 GB memory.}
\label{tab:run-time}
\end{table}

\subsection {MNIST Data Experiment}

In this section we present results of a simple experiment that indicate the potential of generalized resubstitution estimators in image classification by convolutional neural networks (CNN). The experiment uses the well-known MNIST data set and the LeNet-5 CNN architecture. 

The MNIST training data set contains $60000$ $28\times 28$ grayscale images of handwritten digits between 0 and 9 (hence, 10 classes). It is well-known that LeNet-5 can achieve accuracies of upwards of 99\% on this data set; e.g., see \cite{tabik2017snapshot}. 
Problems with small classification error tend to be easier in terms of error estimation performance \cite{BragDougEEPR:15}. To make the problem more challenging, we train the LeNet-5 classifier on random subsets of $n=200, 400, 600$ and $800$ images from the original data set. The remaining data are used to obtain accurate test-set estimates of the true classification error, in order to compute estimates of the bias, variance, and RMS of each error estimator, using 200 independently drawn training data sets for each sample size. The LeNet-5 network was trained using 200 epochs of stochastic gradient descent, with batch size 32, employing 10\% of the training data in each case as a validation data set to stop training early if the validation loss was not reduced for 10 consecutive epochs. 

\begin{figure}
\begin{center}
\begin{tabular}{cccccc}
$\kappa=0$ & \hspace{-0.2in} $\kappa=1$ & \hspace{-0.2in} $\kappa=\kappa^*$ &
$\kappa=0$ & \hspace{-0.2in} $\kappa=1$ & \hspace{-0.2in} $\kappa=\kappa^*$ \\
\includegraphics[scale= 0.25] {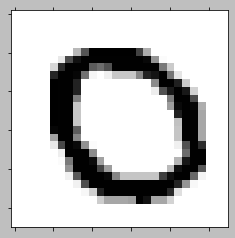}	& 
\hspace{-0.1in}
\includegraphics[scale= 0.25] {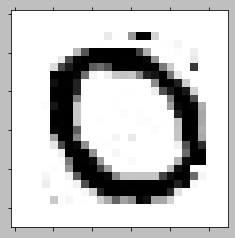}	&
\hspace{-0.1in}
\includegraphics[scale= 0.25] {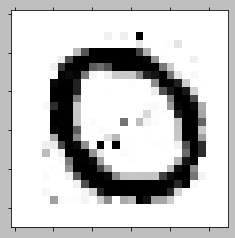}	&
\hspace{-0.1in}
\includegraphics[scale= 0.25] {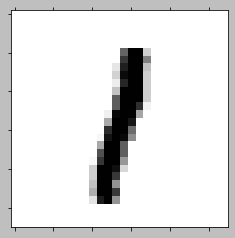}	&
\hspace{-0.1in}
\includegraphics[scale= 0.25] {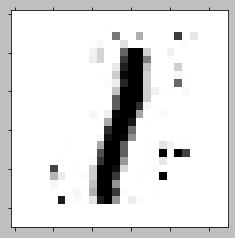}	&
\hspace{-0.1in}
\includegraphics[scale= 0.25] {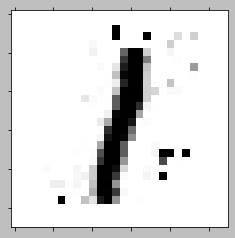}	\\
\includegraphics[scale= 0.25] {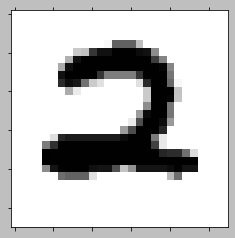}	& 
\hspace{-0.1in}
\includegraphics[scale= 0.25] {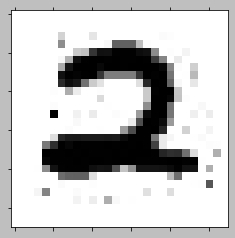}	&
\hspace{-0.1in}
\includegraphics[scale= 0.25] {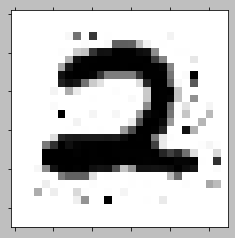}	&
\includegraphics[scale= 0.25] {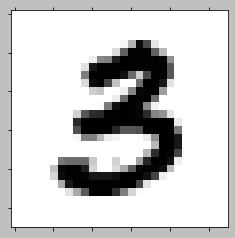}	&
\hspace{-0.1in}
\includegraphics[scale= 0.25] {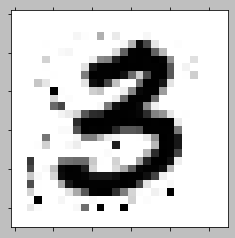}	&
\hspace{-0.1in}
\includegraphics[scale= 0.25] {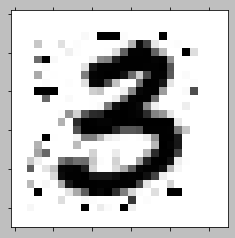}   \\
\includegraphics[scale= 0.25] {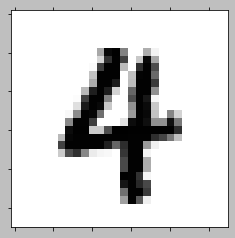}	& 
\hspace{-0.1in}
\includegraphics[scale= 0.25] {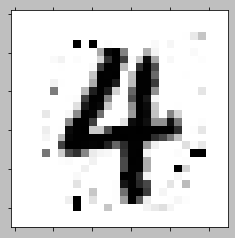}	&
\hspace{-0.1in}
\includegraphics[scale= 0.25] {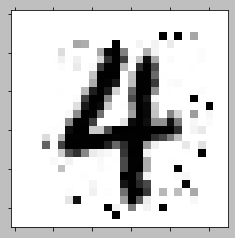}	&
\includegraphics[scale= 0.25] {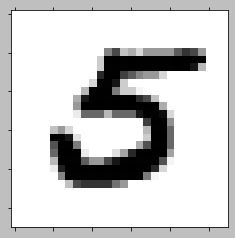}	&
\hspace{-0.1in}
\includegraphics[scale= 0.25] {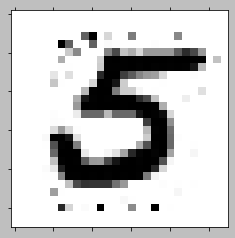}	&
\hspace{-0.1in}
\includegraphics[scale= 0.25] {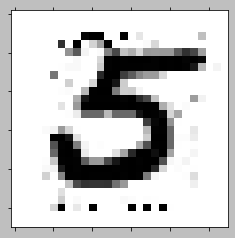}	\\
\includegraphics[scale= 0.25] {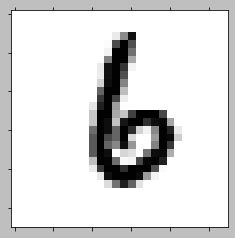}	& 
\hspace{-0.1in}
\includegraphics[scale= 0.25] {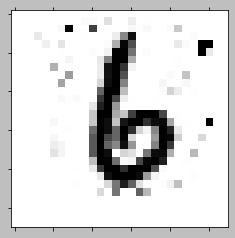}	&
\hspace{-0.1in}
\includegraphics[scale= 0.25] {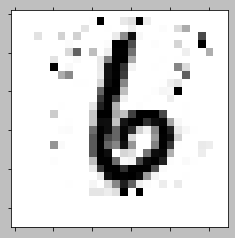}	&
\includegraphics[scale= 0.25] {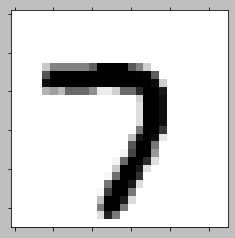}	&
\hspace{-0.1in}
\includegraphics[scale= 0.25] {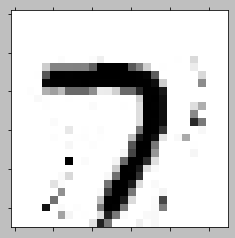}	&
\hspace{-0.1in}
\includegraphics[scale= 0.25] {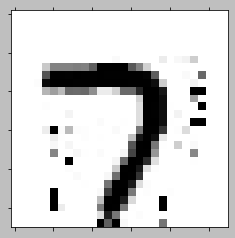}    \\
\includegraphics[scale= 0.25] {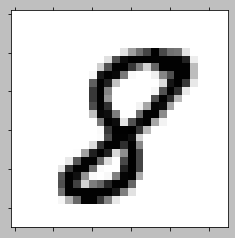}	& 
\hspace{-0.1in}
\includegraphics[scale= 0.25] {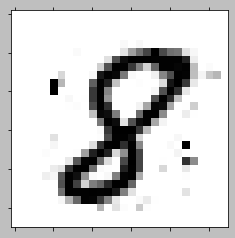}	&
\hspace{-0.1in}
\includegraphics[scale= 0.25] {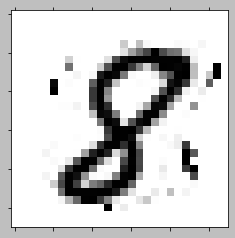}	&
\includegraphics[scale= 0.25] {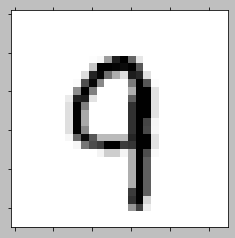}	&
\hspace{-0.1in}
\includegraphics[scale= 0.25] {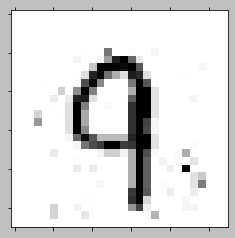}	&
\hspace{-0.1in}
\includegraphics[scale= 0.25] {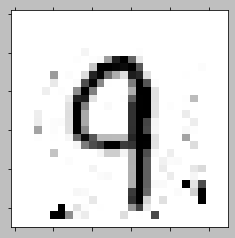}	
\end{tabular}
\end{center}  
    \caption{A few Monte-Carlo images used in the Naive-Bayes bolstered resubstitution error estimator, with $n=600$. The parameter $\kappa^*$ is the optimal correction factor for the calibrated naive Bayes bolstered error estimator (r=1). The cases $\kappa=0$ and $\kappa=1$ refer to the original image and the uncorrected Naive-Bayes bolstered image, respectively.  
    }
    \label{fig:mc_mnist}
\end{figure}

Given that there are 10 classes, the number of neighbors in the KNN-based generalized resubstitution has to increase accordingly; here, we use $k=11$. For bolstering, we  employ diagonal Gaussian kernels, which leads to a ``Naive-Bayes'' bolstering resubstitution estimator, as explained in Section~\ref{Sec-BRE}.
This is done since spherical kernels tend to perform poorly in very high-dimensional spaces~\cite{Simaetal:11}. If $X_{ijk}$ denotes pixel $k$ in image $i$ of class $j$, the mean minimum distance $d_{n,jk}$ among pixels $k$ in class $j$ is:
\begin{equation*}
\hat{d}_{n,jk} \,=\, \frac{1}{n_j} \,\sum_{i=1}^{n_j}\, |X_{ijk}-X_{ijk}^\prime|\,, \quad j=0,1,\ldots,c-1\,,\:k=1,\ldots,28\times 28\,,
\label{eq-dnjk}
\end{equation*}
where $n_j$ is the number of images in class $j$ ($n_j\geq 2$ is assumed) and  $X_{ijk}^\prime$ is the nearest pixel (in value) in position $k$ to $X_{ijk}$ among images in class $j$. The bolstering kernel standard deviations are then given~by:
\begin{equation}
   \sigma_{n,jk} \,=\,\frac{\hat{d}_{n,jk}}{\alpha_{1}}\,, \quad j=0,1,\ldots,c-1\,,\:k=1,\ldots,28\times 28\,,
\label{bolster_sigmaMNIST}
\end{equation}
where $\alpha_1 = 0.674$, as seen in Section~\ref{Sec-BRE}. The naive-Bayes bolstered resubstitution estimator is computed by Monte-Carlo, as in (\ref{er:blstr-mc}), repeated below for convenience:
\begin{equation*}
  \heps^{\,br}_n \,\approx\, \frac{1}{nM} \sum_{i=1}^{n}
     \sum_{j=1}^M  I(\psi_n(\bX_{ij}^{\rm MC})\neq Y_i)\,,
\end{equation*}
where $\{\bX_{ij}^{\rm MC}; j=1,\ldots ,M\}$ are random images generated by drawing each pixel from a Gaussian distribution with mean equal to the original pixel value and standard deviation in~(\ref{bolster_sigmaMNIST}). This generates ``noisy images'' where the intensity of the noise in each pixel is correlated with the variability of pixel values at that position across the training data (for that digit class). Here we employed $M=100$ Monte-Carlo images for each training image. A few of these Monte-Carlo images can be seen in~Figure \ref{fig:mc_mnist}, where the correction factor $\kappa$ is explained later in this section.

\begin{figure}
\begin{center}
\begin{tabular}{ccc}
\includegraphics[scale= 0.345] {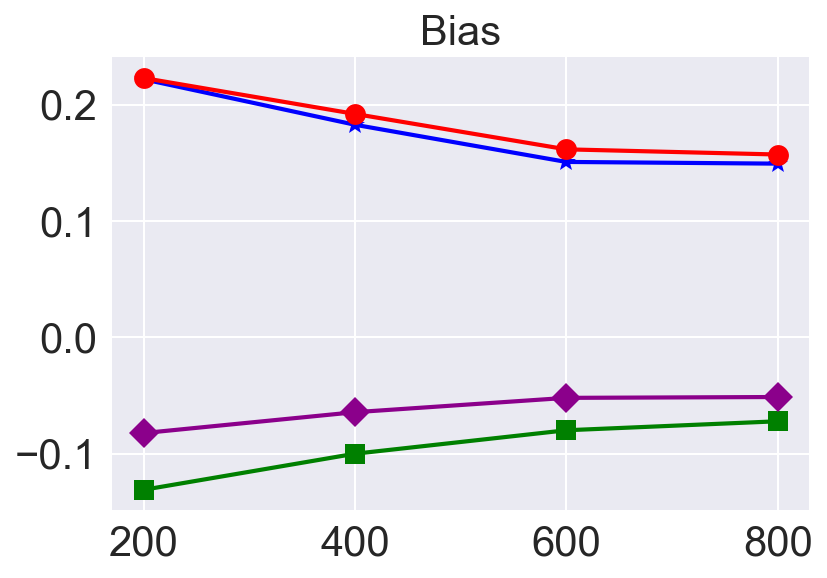}	&  
    \hspace{-0.2in}
	  \includegraphics[scale= 0.345] {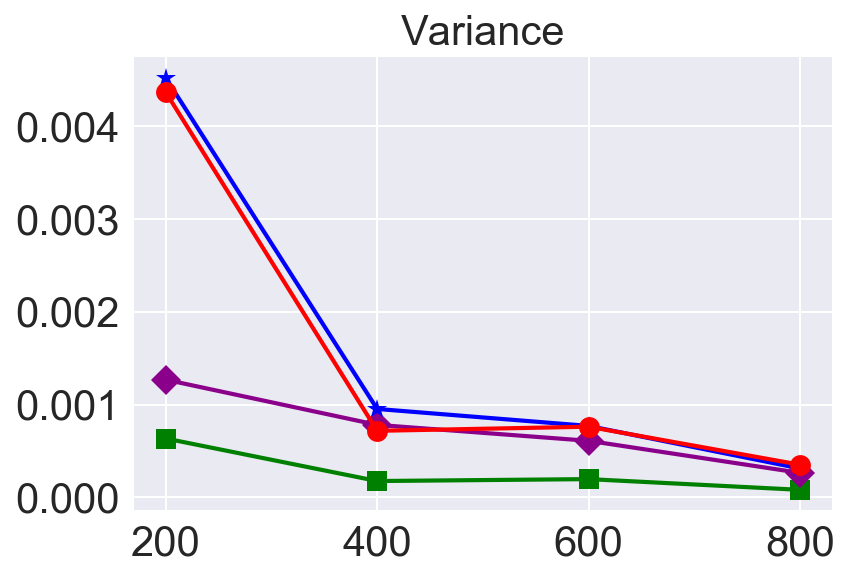} & 
	  \hspace{-0.2in}
	  \includegraphics[scale= 0.345] {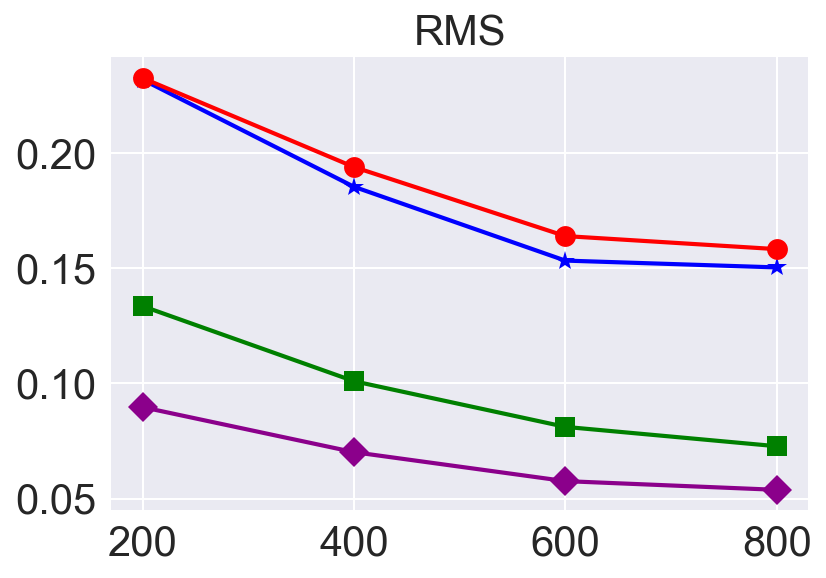} 
\end{tabular}	
\includegraphics[scale=0.4] {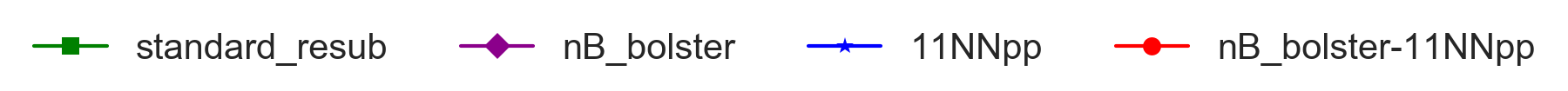}
\end{center}  
    \caption{Bias, variance, and RMS of the deviation distribution for select error estimators on the MNIST data set with LeNet-5 classifier.
    }
    \label{fig:lenet5}
\end{figure}

The results of the experiment are displayed in Figure \ref{fig:lenet5}. We can see that Naive-Bayes bolstered resubstitution is able to improve on the optimistic bias of resubstitution, whereas the 11-NN posterior-probability and Naive-Bayes bolstered posterior-probability error estimators overcompensated and are positively biased, with a larger bias in magnitude as well. None of the generalized resubstitution estimators are able to match the low variance of plain resubstitution, but the Naive-Bayes bolstered resubstitution displays smaller variance than the others at the small sample size $n=200$. When bias and variance are combined in the RMS metric, Naive-Bayes bolstered resubstitution turns out to be the best estimator overall. Notice that the bias, variance, and RMS of all estimators decrease monotonically with increasing sample size.

In order to further reduce the bias of the Naive-Bayes bolstered resubstitution estimator in this high-dimensional space, we employ a data-driven calibration process, which is similar to, but not the same, as the one in \cite{Simaetal:11} (the latter is model-based). We multiply the kernel standard deviation by a constant $\kappa>0$, which is adjusted so as to minimize the estimated bias of the estimator. The bias is roughly estimated by training the classifier on a random sample of 80\% of the images form the available training data, and testing it on remaining 20\%. Depending on the computational cost of training the classifiers, this process can be repeated a number of times $r$ and the results averaged. The point is that the bias does not to be accurately estimated in order to find a useful value for $\kappa$. The calibration process consists of starting at $\kappa=1$, computing the corrected Naive-Bayes bolstered resubstitution estimate, and increasing $\kappa$ by a fixed step-size (here, $0.1$) until the magnitude of the roughly estimated bias does not decrease for two consecutive iterations.

\begin{figure}
\begin{center}
\begin{tabular}{ccc}
\includegraphics[scale= 0.345] {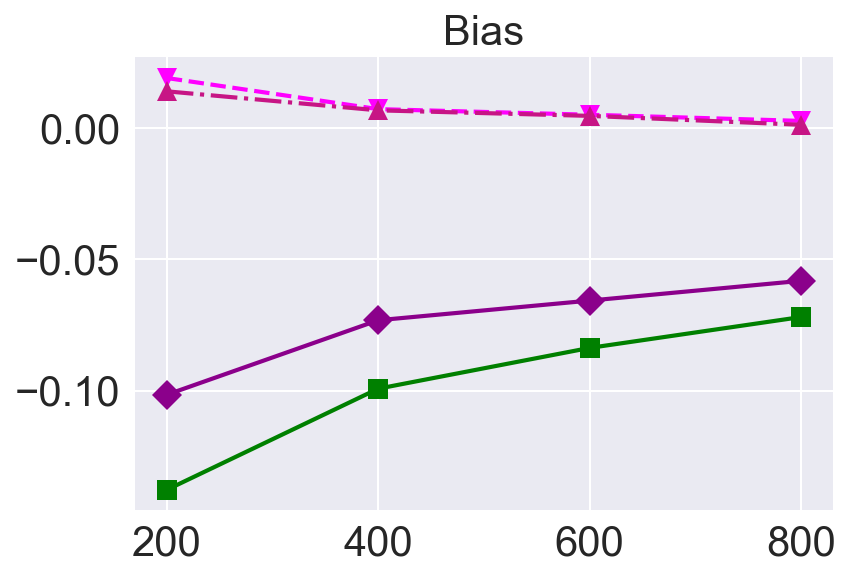}	&  
    \hspace{-0.2in}
	  \includegraphics[scale= 0.345] {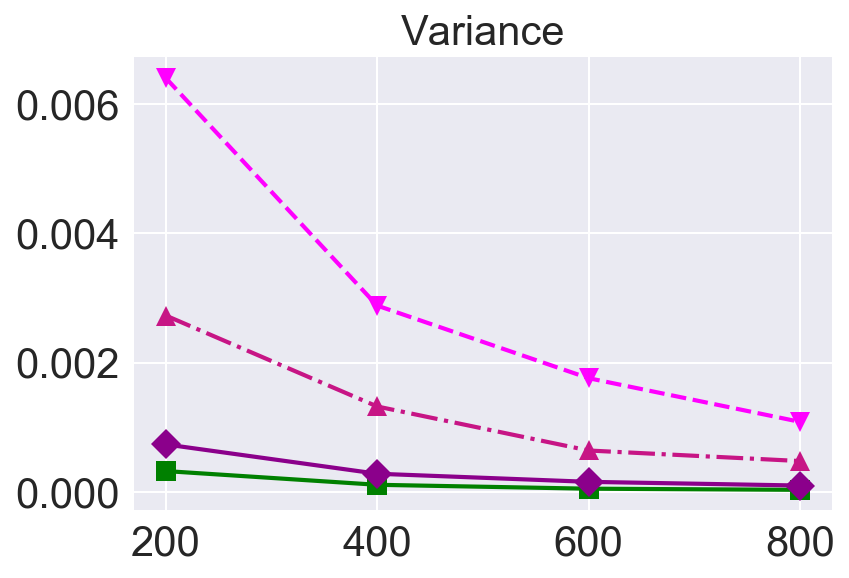} & 
	  \hspace{-0.2in}
	  \includegraphics[scale= 0.345] {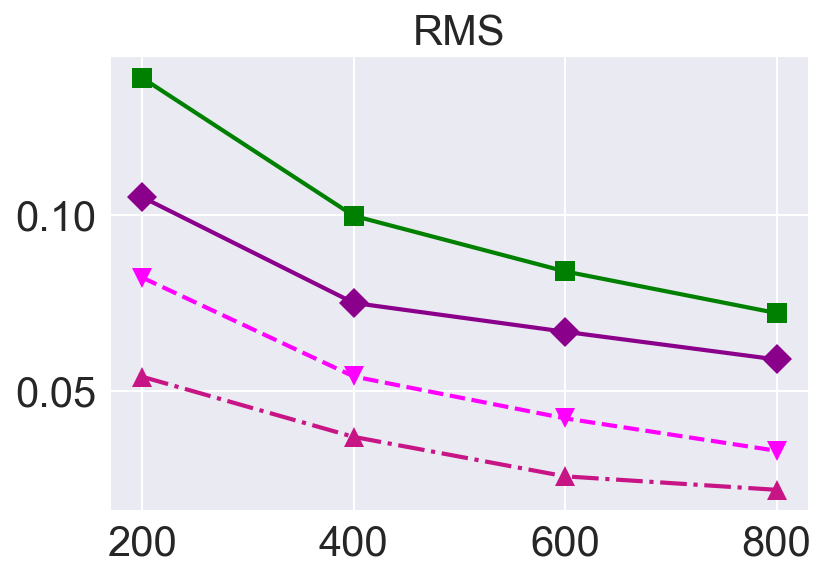} 
\end{tabular}	
\includegraphics[scale=0.4] {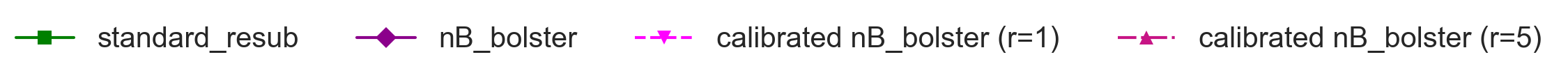}
\end{center}  
    \caption{Calibration of the Naive-Bayes bolstered resubstitution error estimator for the MNIST data.}
    \label{fig:calibrated-semibol}
\end{figure}

Figure \ref{fig:calibrated-semibol} displays the bias, variance, and RMS, computed on the same 200 training data sets as before, of plain resubstitution and Naive-Bayes bolstered resubstitution estimators (these are the same plots as in Figure~\ref{fig:lenet5}), along with the calibrated Naive-Bayes with $r=1$ and $r=5$ repetitions of the rough bias estimation procedure. We can see that calibration succeeded into reducing the bias to nearly zero. However, it did increase the variance, as might be expected (though much less in the case $r=5$ than in the case $r=1$). The RMS curves indicate that the bias-variance tradeoff is in favor of the calibrated estimators, as they achieve significantly smaller RMS than plain Naive-Bayes bolstered resubstitution. Even at $r=1$, (i.e., just one additional step of classifier training), there is a substantial improvement. If more than $r=5$ repetitions are used, at a higher computational cost, it is expected that results will improve further, though a point of diminishing returns will occur eventually.


\section{Conclusions}

The family of generalized resubstitution error estimators was introduced and investigated in this paper. This is a broad family of classification error estimators who can all be computed in the same way by using different empirical measures. They do not require resampling and retraining of classifiers (though in the MNIST classification example a data-driven calibration procedure to further reduce bias was used, which may employ resampling, though it is not necessary). As such, these are generally fast error estimators, which can be used in settings where computational complexity issue is an issue, such as in wrapper feature selection for large data sets. In the two-class case, we showed that these estimators have good large-sample properties, provided that the classification rule has a a finite VC dimension and the corresponding empirical measure converges to the standard empirical measure, in a precise sense. We showed empirically, by means of numerical experiments, that generalized resubstitution error estimators also display excellent small-sample performance. We provided a simple example of application to image classification using deep convolutional neural networks, which indicate the potential of this approach in the area of computer vision, a topic that will be further explored in future work.

\section*{Appendix: Background on Error Estimation}

The subject of classification error estimation has a long history and has
produced a large body of literature; four main review papers summarize
major advances in the field up to 2000
\cite{Tous:74,Hand:86,McLa:87,SchiHand:00}; recent advances in error
estimation since 2000 include work on model selection
\cite{BartBoucLugo:02}, bolstered error estimation
\cite{BragDoug:04a,SimaBragDoug:05}, feature selection
\cite{Simaetal:05,ZhouMao:06,Xiaoetal:07,HancHuaDoug:07}, confidence
intervals \cite{KaarLang:05,Kaar:05,Xuetal:06}, model-based
second-order properties \cite{ZollBragDoug:11,ZollBragDoug:12}, and
Bayesian error estimators \cite{DaltDoug:11a,DaltDoug:11b}.
In this section, we provide a brief review of the basic concepts related to error estimation. A booklength treatment of the topic is provided in \cite{braga2015error}; see also \cite{McLa:92,DGL:96}.

If one knew the distribution of the features and label, then one
could in principle compute the classification error $\eps_n$ by evaluating (\ref{eq:err2}). In practice, such knowledge is rarely available, so
one employs an {\em error estimation rule} $\Xi_n$ in order to obtain a classification error estimate 
\beq
  \hat{\eps}_n \,=\, \Xi_n(\Psi _{n},S_{n},\xi)\,,
\label{eq:errrule}
\eeq
where $\xi$ denote {\em internal random factors} (if any) that
represent randomness that is not introduced by the training data $S_n$; if there are no such internal random factors, the error
estimation rule is said to be {\em \index{error estimation rule!nonrandomized}nonramdomized}, in which case the error estimate is a determined by the data, otherwise, it is
said to be {\em \index{error estimation rule!randomized}randomized}, in which case the error estimator is a random variable given the data. The {\em resubstitution rule} $\Xi _{n}^{\,r}$ is an example of nonrandomized error estimation rule:
\begin{equation*}
  \Xi _{n}^{\,r}(\Psi_{n},S_{n})\,=\,\frac{1}{n}\sum_{i=1}^{n}I(\Psi
_{n}(S_n)(\bX_i)-Y_{i})\,.
\label{eq3-resub}
\end{equation*}
Note that the dependence of $\hat{\eps}_n$ on $\Psi_n$ in (\ref{eq:errrule}) makes explicit the fact that, while the error estimation rule $\Xi_n$ may be fixed (e.g., resubstitution), the properties of $\hat{\eps}_n$ change for different classification rules. This allows us to speak of a resubstitution error estimator $\hat{\eps}^{\,r}_n \,=\, \Xi _{n}^{\,r}(\Psi_{n},S_{n})$ for each classification rule $\Psi_n$. 

\begin{figure}[t]
\begin{center}
\includegraphics[width=0.6\textwidth]{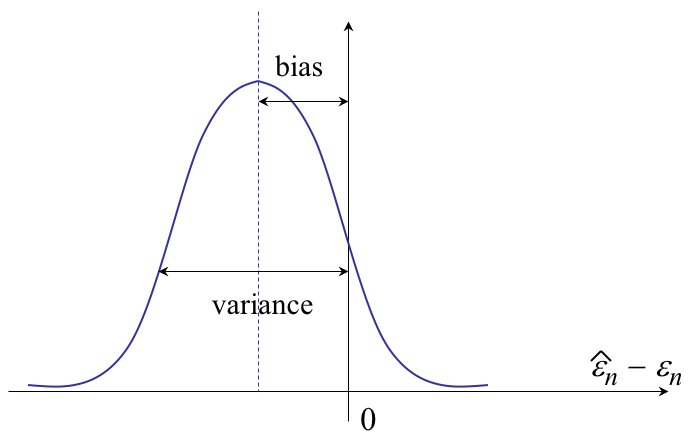}
\end{center}
\par
\vspace{-3ex}
\caption{Deviation distribution showing bias and variance. The
  error estimator in this example is optimistically biased.}
\label{Fig-devdist}
\end{figure}

The performance of an error estimator can be assessed by the distribution of $\hat{\eps}_n - \eps_n$, called the \emph{deviation distribution}~\cite{BragDoug:04}. For good performance, this distribution should be peaked (low-variance) and centered near zero (low-bias). 
See Figure~\ref{Fig-devdist} for an illustration. 

The {\em bias} is defined as the first moment of the deviation distribution:
\begin{equation}
\mathrm{Bias}(\hat{\varepsilon}_{n})\,=\,E[\hat{\varepsilon}_{n}-\varepsilon
_{n}]\,=\,E[\hat{\varepsilon}_{n}]-E[\varepsilon _{n}]\,.  
\label{eq-bias}
\end{equation}
The error estimator $\hat{\varepsilon}_{n}$ is said to be
\emph{\index{error estimator!optimistically biased}optimistically biased} if
$\mathrm{Bias}(\hat{\varepsilon}_{n})<0$ and
\emph{\index{error estimator!pessimistically biased}pessimistically biased} if
$\mathrm{Bias}(\hat{\varepsilon}_{n})>0$. It is {\em \index{error estimator!unbiased}unbiased}
if $\mathrm{Bias}(\hat{  \varepsilon}_{n})=0$. The resubstitution error estimator is usually
optimistically biased.

The {\em deviation variance} is the variance of the deviation distribution:
\vspace{-0.5ex} 
\begin{equation}
\mathrm{Var}_{\mathrm{dev}}(\hat{\varepsilon}_{n})\,=\,\mathrm{Var}(\hat{\varepsilon}_{n}-\varepsilon _{n})\,=\, \mathrm{Var}(\hat{\varepsilon}_{n})+\mathrm{Var}(\varepsilon _{n}) -2\mathrm{Cov}(\varepsilon _{n},\hat{\varepsilon}_{n})\,.
\label{eq-err_vd}
\end{equation}
Unlike in classical statistics, where estimators for fixed parameters
are sought, here the quantity being estimated, namely
$\varepsilon_{n}$, is random and thus a ``moving target.''
This is why it is appropriate to consider the variance of the
difference, $\var(\hat{\varepsilon}_{n}-\varepsilon_{n})$. However,
if the classification rule is not overfitting, then $\var(\varepsilon_n) \approx 0$ --- in fact, overfitting could be
defined as present if $\var(\varepsilon_n)$ is large, since in that
case the classification rule is learning the changing data and not the
fixed underlying feature-label distribution. It follows, from the Cauchy-Schwartz
Inequality that
$\cov(\varepsilon _{n},\hat{\varepsilon}_{n}) \leq
\sqrt{\var(\varepsilon _{n})\var(\hat{\varepsilon}_{n})} \approx 0$,
and thus, from (\ref{eq-err_vd}), $\var(\hat{\varepsilon}_{n}-\varepsilon _{n}) \approx
\var(\hat{\varepsilon}_{n})$. If an estimator is randomized, then
it has additional {\em internal variance}
$V_{\mathrm{int}}\,=\,\var(\hat{\varepsilon}_{n}|S_{n})$, which 
measures the variability due only to the internal random
factors, while the full variance $\var(\hat{\varepsilon}_{n})$
measures the variability due to both the sample $S_{n}$ and the
internal random factors~$\xi$. The following formula can be easily shown using the Conditional Variance
Formula of probability theory:
\begin{equation}
  \var(\hat{\varepsilon}_{n})\,=\,E[V_{\mathrm{int}}] + 
 \var(E[\hat{\varepsilon}_{n}|S_{n}])\,.
\label{eq-var_est}
\end{equation}
The first term on the right-hand side contains the contribution of the internal
variance to the total variance. For nonrandomized
$\hat{\varepsilon}_{n}$, $V_{\mathrm{int}}=0$; for randomized
$\hat{\varepsilon}_{n}$, $E[V_{\mathrm{int}}] > 0$. 

The \emph{root mean-square error} is the square root of the
second moment of the deviation distribution:
\begin{equation}
\mathrm{RMS}(\hat{\varepsilon}_{n})\,=\,\sqrt{E[(\hat{\varepsilon}_{n}-\varepsilon _{n})^{2}]}\,=\,\sqrt{\mathrm{Bias}(\hat{\varepsilon}_{n})^{2}+\mathrm{Var}_{\mathrm{dev}}(\hat{\varepsilon}_{n})}
\label{eq-err_rms}
\end{equation}
The RMS is generally considered the most important error estimation
performance metric.  The other performance metrics appear within the
computation of the RMS; indeed, all of the five basic moments --- the
expectations $E[\varepsilon _{n}]$ and $E[\hat{\varepsilon}_{n}]$, the variances $\mathrm{Var}(\varepsilon_{n})$ and $\mathrm{Var}(\hat{\varepsilon}_{n})$, and the covariance
$\mathrm{Cov}(\varepsilon _{n},\hat{\varepsilon}_{n})$ --- appear within the
RMS. 

Finally, an error estimator is said to be \emph{\index{error estimator!consistent}consistent} if $\hat{\varepsilon}_{n}\rightarrow
\varepsilon _{n}$ in probability as $n\rightarrow \infty $, and
\emph{\index{error estimator!strongly consistent}strongly consistent} if convergence is with
probability $1$. By an application of Markov's Inequality, we have
\begin{equation}
P(|\hat{\varepsilon}_{n}-\varepsilon _{n}|\geq \tau) = P(|\hat{\varepsilon}_{n}-\varepsilon_{n}|^2\geq \tau^2)\,\leq\, \frac{E[|\hat{\varepsilon}_{n}-\varepsilon _{n}|^2]}{\tau ^{2}} \,=\,\left(\frac{\mathrm{RMS}(\hat{\varepsilon}_{n})}{\tau}\right)^{2},\quad \text{for }\tau >0\,.  \label{eq-RMS_tail}
\end{equation}
Hence, if 
$\mathrm{RMS}(\hat{\varepsilon}_{n}) \rt 0$, then $P(|\hat{\varepsilon}_{n}-\varepsilon _{n}|\geq \tau ) \rt 0$, for any $\tau>0$, i.e. the error estimator is consistent.

Good error estimation performance requires that the bias,
deviation variance, and RMS be as close as
possible to zero.

\small
\bibliographystyle{elsarticle-num}
\bibliography{refs.bib,all.bib}

\end{document}